\documentclass{article}

\usepackage[final]{neurips_2025}

\usepackage[utf8]{inputenc} 
\usepackage[T1]{fontenc}    
\usepackage{hyperref}       
\usepackage{url}            
\usepackage{booktabs}       
\usepackage{amsfonts}       
\usepackage{amsmath}        
\usepackage{nicefrac}       
\usepackage{microtype}      
\usepackage{xcolor}         
\usepackage{graphicx}
\usepackage{algorithmic}
\usepackage[linesnumbered, vlined, ruled]{algorithm2e}
\newcommand{\RNum}[1]{\uppercase\expandafter{\romannumeral #1\relax}}
\usepackage{amssymb} 
\usepackage{geometry} 
\usepackage{bm} 
\usepackage{enumitem} 
\usepackage{subfig}

\usepackage{nicematrix}
\usepackage{booktabs}

\usepackage{caption}
\usepackage{comment}

\title{PhysVarMix: Physics-Informed Variational Mixture Model for Multi-Modal Trajectory Prediction}

\author{%
  Haichuan Li \\
  Turku Intelligent Embedded and Robotics Systems lab, Faculty of Technology\\
  University of Turku\\
  Turku, Finland \\
  \texttt{haicli@utu.fi} \\
  \And
  Tomi Westerlund \\
  Turku Intelligent Embedded and Robotics Systems lab, Faculty of Technology \\
  University of Turku\\
  Turku, Finland \\
  \texttt{tovewe@utu.fi} \\
}

\begin{document}

\maketitle

\begin{abstract}

Accurate prediction of future agent trajectories is a critical challenge for ensuring safe and efficient autonomous navigation, particularly in complex urban environments characterized by multiple plausible future scenarios. In this paper, we present a novel hybrid approach that integrates learning-based with physics-based constraints to address the multi-modality inherent in trajectory prediction. Our method employs a variational Bayesian mixture model to effectively capture the diverse range of potential future behaviors, moving beyond traditional unimodal assumptions. Unlike prior approaches that predominantly treat trajectory prediction as a data-driven regression task, our framework incorporates physical realism through sector-specific boundary conditions and Model Predictive Control (MPC)-based smoothing. These constraints ensure that predicted trajectories are not only data-consistent but also physically plausible, adhering to kinematic and dynamic principles. Furthermore, our method produces interpretable and diverse trajectory predictions, enabling enhanced downstream decision-making and planning in autonomous driving systems. We evaluate our approach on two benchmark datasets, demonstrating superior performance compared to existing methods. Comprehensive ablation studies validate the contributions of each component and highlight their synergistic impact on prediction accuracy and reliability. By balancing data-driven insights with physics-informed constraints, our approach offers a robust and scalable solution for navigating the uncertainties of real-world urban environments.

\end{abstract}

\section{Introduction}
	
Trajectory prediction is a critical component of autonomous driving systems, enabling vehicles to anticipate the future positions of surrounding agents and plan their own motion accordingly~\cite{leon2021review, gomes2022review}. Despite significant advances in perception and control algorithms, accurate prediction remains challenging due to the complex, multi-agent nature of traffic scenarios and the inherent uncertainty in human behavior~\cite{huang2022survey, karunakaran2023efficient}. In urban environments, where multiple road users interact and various path options exist, the future trajectory of an agent is inherently multi-modal—a vehicle approaching an intersection, for instance, might turn left, right, or continue straight with varying probabilities that depend on numerous factors including road geometry, traffic conditions, and agent history~\cite{mozaffari2022multimodal, gilles2021probabilistic}.
Traditional approaches to trajectory prediction often rely on handcrafted kinematic models~\cite{lefevre2014survey, schreier2016integrated} or data-driven methods using neural networks~\cite{mozaffari2021deep, jeong2019motion}. While kinematic models provide physically consistent predictions, they struggle to capture the diversity of possible futures. Conversely, pure learning-based approaches can model complex relationships but may generate physically implausible trajectories that violate kinematic constraints~\cite{zhao2021review, mercat2020multi}. More recent methods have employed generative models~\cite{huang2022survey, salzmann2020trajectron} and graph neural networks~\cite{li2021spatio} to address the multi-modality challenge, but many still lack the ability to produce diverse yet feasible predictions within a unified framework. Rule-based approaches have achieved some success in industry but require extensive human engineering to deal with diverse real-world scenarios~\cite{ulbrich2015towards}.
The integration of map information presents another significant challenge. Road networks impose strong priors on vehicle movement, yet encoding this structural information in a way that neural networks can effectively utilize remains difficult~\cite{wang2023survey}. While some approaches represent map features as rasterized images~\cite{chen2022learning}, others employ vectorized representations~\cite{zhou2023vector}, each with their own limitations in capturing the hierarchical relationships between agents and the environment.

In this paper, we propose a method that addresses these challenges through a principled approach to trajectory prediction. Our algorithm combines the representational capacity of a data-driven strategy with the probabilistic framework of variational inference and the physical consistency of model predictive control. The key contributions of our work are:
\begin{itemize}
    \item \textbf{Hierarchical Scene Encoding}: We introduce a two-level encoding architecture that first processes individual polylines (lanes, crosswalks, agent histories) locally before performing global cross-attention across different feature types, effectively capturing both fine-grained details and high-level scene context.
    \item \textbf{Causal-based Mask}: captures temporal dependencies and agent-environment interactions, maintaining state embeddings across prediction steps to ensure temporal consistency.
    \item \textbf{Variational Bayesian Mixture Model}: We develop a variational prediction head that explicitly models trajectory distributions as a mixture of Gaussians, capturing the inherent multi-modality of future behaviors while quantifying uncertainty in a principled manner.
    \item \textbf{Physics-Informed Constraints}: We incorporate domain knowledge through sector boundary constraints and an MPC-based trajectory smoother that ensures kinematic feasibility without sacrificing predictive accuracy or diversity.
\end{itemize}

\begin{figure}[t]
    \centering
    \includegraphics[width=\textwidth]{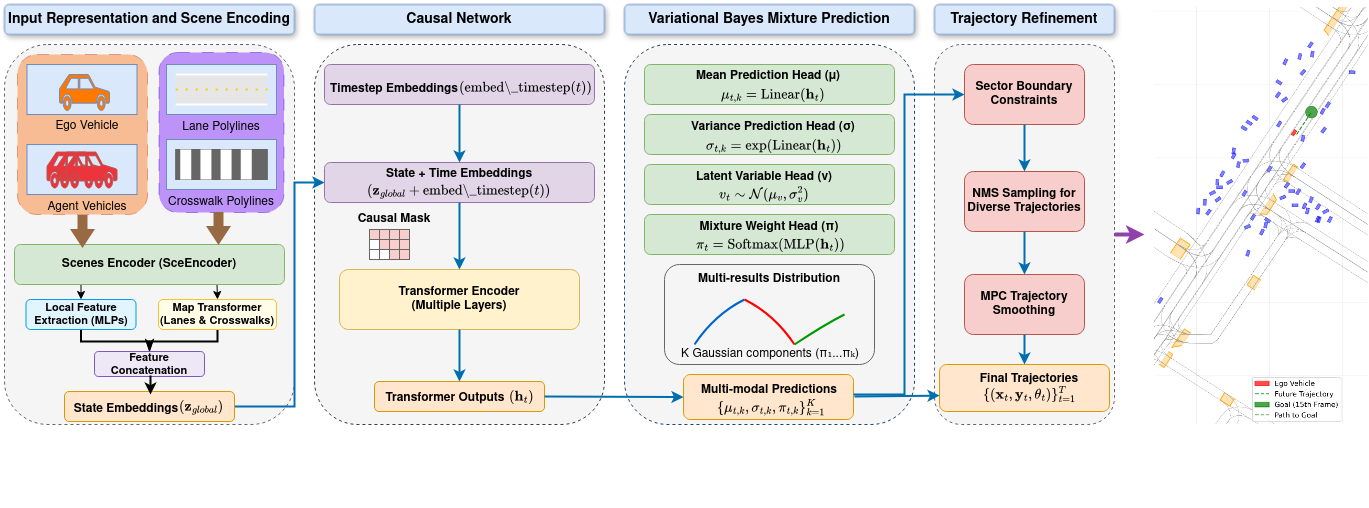}
    \caption{Overview of the \textit{PhysVarMix} framework. 
Model Process Flow:
\RNum{1:} Scene encoders process scene elements (ego vehicle, agents, map)
\RNum{2:} Causal Network integrates historical and current scenarios
\RNum{3:} The variational mixture model predicts multi-modal trajectory distributions
\RNum{4:} Refinement steps apply physical constraints and optimize trajectories.}
    \label{fig:overview}
\end{figure}

\section{Related Work}

Trajectory prediction for autonomous vehicles has evolved from deterministic forecasting to probabilistic, multimodal approaches that better capture the inherent uncertainty of future motion. Early methods relied on physics-based models or simple recurrent neural networks (RNNs)~\cite{hochreiter1997long} to extrapolate agent motion, but these struggled with the diversity and complexity of real-world scenarios~\cite{mozaffari2021deep}. Mixture Density Networks (MDNs)~\cite{bishop1994mixture} and Conditional Variational Autoencoders (CVAEs)~\cite{sohn2015learning} introduced the concept of modeling future trajectories as distributions, enabling the generation of multiple plausible futures. Earlier methods such as MultiPath~\cite{chai2019multipath}, MTP~\cite{cui2018multimodal}, and LaneGCN~\cite{liang2020learning} have advanced multimodal prediction by leveraging scene context and agent interactions, though they often face challenges like mode collapse or redundant trajectories. 

Recent advancements include innovative approaches such as~\cite{bae2024language}, who proposed LMTraj, a language-based multimodal trajectory predictor that recasts the prediction task as a question-answering problem, achieving competitive performance. Additionally,~\cite{sriramulu2024multitransmotion} introduced Multi-Transmotion, a pre-trained model for human motion prediction that integrates multiple datasets and employs a transformer-based architecture~\cite{vaswani2017attention} for cross-modality pre-training, enhancing the robustness of multimodal predictions.

The transformer architecture~\cite{vaswani2017attention} has become a cornerstone of sequence modeling tasks, including trajectory prediction, due to its ability to capture long-range dependencies and complex agent-environment interactions through self-attention mechanisms. Models like Scene Transformer~\cite{ngiam2021scene}, Wayformer~\cite{nayakanti2022wayformer}, and AutoBots~\cite{girgis2022latent} have demonstrated the effectiveness of transformers in modeling temporal and spatial relationships in multi-agent scenarios. However, these models often lack explicit enforcement of causality or physical feasibility, which can result in temporally inconsistent or physically implausible predictions. 

Recent research has addressed these limitations by enhancing transformer-based approaches. For instance,~\cite{zhou2024smartpretrain} proposed SmartPretrain, a model-agnostic and dataset-agnostic representation learning method that improves generalization across diverse scenarios. Similarly,~\cite{leng2024trajllm} explored the integration of pre-trained large language models to enhance trajectory prediction, leveraging their advanced contextual understanding capabilities.

Ensuring both diversity and feasibility in predicted trajectories remains a significant challenge. Techniques such as Non-Maximum Suppression (NMS), originally developed for object detection~\cite{ren2015faster}, have been adapted to filter redundant trajectory samples~\cite{lee2017desire}. Other approaches, such as DESIRE~\cite{lee2017desire} and DLow~\cite{yuan2020dlow}, employ ranking, refinement, or adversarial objectives to promote diversity, though these often require additional training objectives or post-processing steps. Physical feasibility is typically addressed through kinematic constraints, physics-based loss terms, or post-hoc filtering, but these solutions may not be seamlessly integrated with the prediction model. Recent advancements have leveraged generative models to address these issues. For example,~\cite{gao2024intentiondiffusion} introduced an intention-aware denoising diffusion model that generates diverse trajectories by considering the agent's intention. Similarly,~\cite{kim2024singulartrajectory} proposed SingularTrajectory, a universal trajectory predictor that uses a diffusion model to ensure diverse and feasible predictions.
Recent research has emphasized the integration of physical constraints and causal reasoning to enhance the safety and reliability of trajectory prediction. Physics-guided neural networks, such as those proposed by~\cite{tumu2023physics}, incorporate physical models to restrict predictions to dynamically feasible regions, using a surrogate dynamical model to ensure trajectory feasibility. Similarly,~\cite{li2025hybrid} developed a hybrid machine learning model that combines deep learning with kinematic motion models, enforcing physical feasibility through constrained action spaces. On the causal reasoning front,~\cite{xia2024robusttrajectory} employed causal learning to isolate environmental confounders, enhancing the robustness of trajectory representations. Additionally,~\cite{hari2024navigation} explored trajectory prediction under uncertainty using switching dynamical systems, which incorporate occlusion reasoning and maintain uncertainty about occluded objects, further improving prediction reliability.

\section{Methods}
This section elucidates the proposed framework for precise trajectory prediction, designed to integrate heterogeneous data sources for robust forecasting. We utilize intermediate input representations to improve generalization and interoperability, rather than directly learning an end-to-end policy from raw sensor data, such as camera images or lidar point clouds. The methodology is systematically delineated, beginning with an overview of the model architecture, followed by a detailed exposition of each constituent component.
: the \textit{Scene encoder}, \textit{causal-based network}, \textit{variational mixture prediction head}, \textit{physics-informed constraints}, and \textit{trajectory optimization module}.

\subsection{Model Architecture Overview}
The proposed model adopts a multi-stage architecture, augmented with probabilistic modeling and physics-informed constraints to enhance prediction accuracy and physical plausibility. As depicted in Figure~\ref{fig:overview}, the framework comprises five primary components:
\begin{itemize}
    \item \textbf{Scene Encoder}: Processes diverse inputs, including the ego vehicle's historical states, trajectories of surrounding agents, and contextual map elements such as lanes and crosswalks.
    \item \textbf{Causal Network}: Captures temporal dependencies across multiple time steps, enabling the modeling of dynamic interactions.
    \item \textbf{Variational Bayesian Mixture Prediction Head}: Generates a distribution of diverse trajectory predictions, accounting for multi-modal motion possibilities.
    \item \textbf{Physics-Informed Constraints}: Enforces kinematic and dynamic feasibility, ensuring that predicted trajectories adhere to physical limitations.
    \item \textbf{Model Predictive Control (MPC) Smoother}: Optimizes predicted trajectories to guarantee dynamical consistency and smoothness.
\end{itemize}

Given the historical states of the ego vehicle, surrounding agents, and contextual map data, our algorithm predicts a set of trajectories, each representing a possible motion. This enumeration approach facilitates comprehensive scene analysis and robust decision-making in autonomous systems.

\begin{figure}[h]
    \centering
    \includegraphics[width=\textwidth]{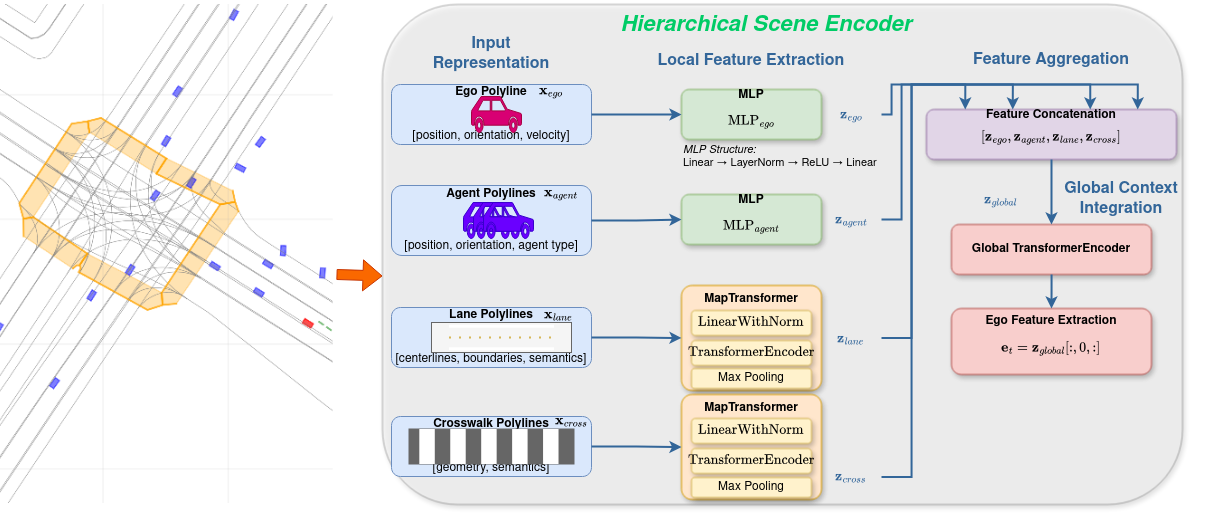}
    \caption{Overview of the \textit{PhysVarMix} framework. 
Model Process Flow:
1. Scene encoders process scene elements (ego vehicle, agents, map)
2. Causal transformer integrates historical and current scenarios
3. Variational mixture model predicts multi-modal trajectory distributions
4. Refinement steps apply physical constraints and optimize trajectories}
    \label{fig:encoder}
\end{figure}

\subsection{Scene Encoding}

The scene encoder processes raw input consisting of heterogeneous entities (ego vehicle, surrounding agents, lane segments, and crosswalks), each represented as polylines with associated features.

\paragraph{Input Representation}

For each entity type, we construct corresponding polyline features:
\textit{Ego Polyline}: Represents the ego vehicle's historical states, including position, orientation, velocity, and acceleration over the past $T_{hist}$ frames. Moreover, an ego-perturbed goal-oriented coordinate system is applied. The origin of this coordinate system is anchored at the ego vehicle’s position plus a zero-mean Gaussian perturbation. The perturbation blurs the ego vehicle’s position information and ensures local observability. 
\textit{Agent Polylines}: Capture surrounding agents' historical states, including position, orientation, and agent type.
\textit{Lane Polylines}: Encode lane center lines, boundaries, and semantic attributes (e.g., lane type, traffic light state).
\textit{Crosswalk Polylines}: Represent pedestrian crossings with corresponding geometry and semantics.
Each polyline is represented as a sequence of points with associated features, with masks to handle variable numbers of entities.

\paragraph{Hierarchical Encoding}

We employ a hierarchical encoding strategy(Fig.~\ref{fig:encoder}) in our algorithm:

\textit{Local Feature Extraction}: For map elements (lanes and crosswalks), we first apply a MapTransformer that processes each polyline independently:
   $\mathbf{z}_{lane} = \text{MapTransformer}(\mathbf{x}_{lane})$
   $\mathbf{z}_{cross} = \text{MapTransformer}(\mathbf{x}_{cross})$.
The MapTransformer consists of a linear embedding layer followed by an encoder that operates on the points within each polyline, enabling the model to capture local geometric patterns.
\textit{Agent Encoding}: Ego and surrounding agent histories are processed with dedicated MLPs:
   $\mathbf{z}_{ego} = \text{MLP}_{ego}(\mathbf{x}_{ego})$
   $\mathbf{z}_{agent} = \text{MLP}_{agent}(\mathbf{x}_{agent})$.
\textit{Global Context Integration}: The locally processed features are concatenated and fed into a global encoder to model cross-entity relationships:
   $\mathbf{z}_{global} = \text{Encoder}([\mathbf{z}_{ego}, \mathbf{z}_{agent}, \mathbf{z}_{lane}, \mathbf{z}_{cross}])$.

This allows the model to attend between different entity types, capturing interactions between agents and their relationship to the map.
The final output of the scene encoder is a sequence of embedded states for the ego vehicle, denoted as $\mathbf{e}t$.

\begin{figure}[ht]
    \centering
    \subfloat[Causal Mask]{\includegraphics[height=3.5cm]{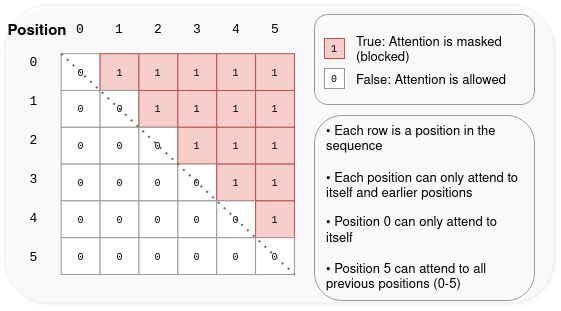}\label{subfig:mask}}
    \hfill
    \subfloat[Sector Boundary Constraints for Trajectory]{\includegraphics[height=3.5cm]{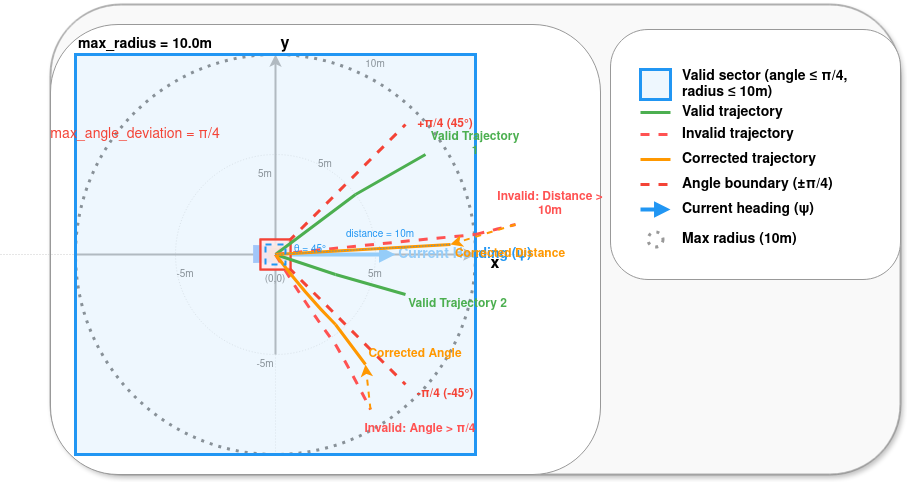}\label{subfig:sector}}
    \caption{(a): The causal mask ensures autoregressive prediction where each timestep can only attend to current and past information, preventing information leakage from the future. (b): Correction Algorithm: \RNum{1}. For points outside angle bounds: $\theta\_corrected = min(max(\theta, \Psi-\pi/4), \Psi+\pi/4)$ \RNum{2}. For points outside distance bounds
    \RNum{3}. Converting back to Cartesian coordinates
    .}
    \label{fig:subfigs}
\end{figure}

\subsection{Causal Network}

To model temporal dependencies, we employ a causal-based network that processes the sequence of encoded scenarios over time. To combine with the temporal features, we use a temporal encoder to embed the spatial scenes features and generate predictions. Given a sequence of input agent states $\{\mathbf{x}_{t-L+1}, \ldots, \mathbf{x}_t\}$, where $L$ is the causal context length, each state is first encoded: $\mathbf{e}\_i = \mathrm{SceneEncoder}(\mathbf{x}\_i), \quad i = t-L+1, \ldots, t$. 
To combine temporal information, we add a learned positional embedding:
$\tilde{\mathbf{e}}_i = \mathbf{e}_i + \mathrm{TimeEmbed}(i)$
where $\mathrm{TimeEmbed}$ is an embedding lookup table.
The sequence $\{\tilde{\mathbf{e}}_{t-L+1}, \ldots, \tilde{\mathbf{e}}_t\}$ is processed by a stack of neural layers with a \textbf{causal mask}(Fig.~\ref{subfig:mask}) $\mathbf{M} \in \mathbb{R}^{L \times L}$:
$\text{mask}[i, j] = \begin{cases}
0 & \text{if } i \geq j \\
-\infty & \text{otherwise}
\end{cases}$,
This mask ensures that the output at time $j$ only attends to inputs at times $i \leq j$, strictly preventing information leakage from the future.
The causal network then computes:
$\mathbf{H} = \mathrm{Transformer}(\tilde{\mathbf{e}}_{t-L+1:t}; \mathbf{M})$
where $\mathbf{H} \in \mathbb{R}^{L \times d}$ is the sequence of hidden states.
During training, the hidden states are aggregated by averaging:
$\mathbf{h} = \frac{1}{L} \sum_{i=1}^{L} \mathbf{H}_i$
During inference, only the last hidden state is used:
$\mathbf{h} = \mathbf{H}_L$
This causal network enables the model to capture temporal dependencies and agent-environment interactions, while strictly enforcing the autoregressive property required for sequential prediction.


\subsection{Variational Mixture Prediction Head}

To capture the multi-modal nature of future trajectories, we employ a variational Bayesian mixture model that predicts parameters for $K$ distinct trajectory modes. Specifically, for each mixture component $k \in \{1, \ldots, K\}$, the model predicts a mean trajectory $\boldsymbol{\mu}_k \in \mathbb{R}^{T_{\text{future}} \times D_{\text{act}}}$, a diagonal covariance $\boldsymbol{\Sigma}_k \in \mathbb{R}^{T_{\text{future}} \times D_{\text{act}}}$, and a mixture weight $\pi_k \in [0, 1]$ such that $\sum_{k=1}^K \pi_k = 1$. These parameters collectively define a mixture of Gaussians over the future trajectory space:
\[
p(\mathbf{y} | \mathbf{h}) = \sum_{k=1}^K \pi_k \, \mathcal{N}(\mathbf{y} \mid \boldsymbol{\mu}_k, \boldsymbol{\Sigma}_k) = \sum_{k=1}^K \pi_k \, \frac{1}{(2\pi)^{D/2} |\boldsymbol{\Sigma}_k|^{1/2}} \exp\left(-\frac{1}{2} (\mathbf{y} - \boldsymbol{\mu}_k)^T \boldsymbol{\Sigma}_k^{-1} (\mathbf{y} - \boldsymbol{\mu}_k)\right)
\]
where $\mathbf{y}$ denotes a future trajectory and $\mathbf{h}$ is the context embedding produced by the causal network.

To further enhance the expressiveness of the model, we introduce a latent variable $\mathbf{v} \in \mathbb{R}^{D_{\text{latent}}}$, which captures additional variations within each mode. The overall predictive distribution is formulated as:
\[
p(\mathbf{y} | \mathbf{h}) = \int p(\mathbf{y} | \mathbf{v}, \mathbf{h}) \, p(\mathbf{v} | \mathbf{h}) \, d\mathbf{v}
\]
Here, $p(\mathbf{v} | \mathbf{h})$ is a diagonal Gaussian, with its parameters predicted from the context embedding:
\[
\boldsymbol{\mu}_v = \mathrm{LSTM}_{v_\mu}(\mathbf{h}), \qquad
\boldsymbol{\sigma}_v = \exp(\mathrm{Linear}_{v_\sigma}(\mathbf{h}))
\]
During training, the latent variable $\mathbf{v}$ is sampled using the reparameterization trick. To model the temporal evolution of the latent variable over the prediction horizon, we employ an additional LSTM
\[
\mathbf{v} = \boldsymbol{\mu}_v + \boldsymbol{\sigma}_v \odot \boldsymbol{\epsilon}, \quad \boldsymbol{\epsilon} \sim \mathcal{N}(\mathbf{0}, \mathbf{I}),
\qquad
\mathbf{v}_{1:T_{\text{future}}} = \mathrm{LSTM}_{v_{\text{trans}}}(\mathbf{v}, T_{\text{future}})
\]
Finally, the mixture parameters are computed by dedicated prediction heads as follows:
\[
\boldsymbol{\mu}_{1:K} = \mathrm{Linear}_\mu(\mathbf{h}), \qquad
\boldsymbol{\sigma}_{1:K} = \exp(\mathrm{Linear}_\sigma(\mathbf{h})), \qquad
\boldsymbol{\pi} = \mathrm{Softmax}(\mathrm{MLP}_\pi(\mathbf{h}))
\]
This design enables the model to flexibly represent complex, multi-modal distributions over future trajectories, while maintaining efficient and stable training dynamics.


\subsection{Physics-Informed Constraints}
To ensure physically plausible predictions, we incorporate domain knowledge through explicit constraints on the predicted trajectories.

\paragraph{Sector Boundary Constraint.}
Vehicles are subject to kinematic limitations on turning radius and acceleration. We implement these limitations via a sector boundary constraint~\ref{subfig:sector}, which restricts predicted trajectories to a feasible region defined by:
\RNum{1}. A maximum radius $r_{\max}$ from the current position.
\RNum{2}. A maximum angular deviation $\theta_{\max}$ from the current heading.

For each predicted point $(\hat{x}, \hat{y})$, we compute the relative distance and angle:
\[
r = \sqrt{(\hat{x} - x_0)^2 + (\hat{y} - y_0)^2}, \qquad
\theta = \mathrm{atan2}(\hat{y} - y_0, \hat{x} - x_0) - \theta_0
\]
If the point lies outside the sector ($r > r_{\max}$ or $|\theta| > \theta_{\max}$), we project it onto the sector boundary:
\[
r' = \min(r, r_{\max}), \qquad
\theta' = \mathrm{clamp}(\theta, -\theta_{\max}, \theta_{\max})
\]
\[
\hat{x}' = x_0 + r' \cos(\theta' + \theta_0), \qquad
\hat{y}' = y_0 + r' \sin(\theta' + \theta_0)
\]
This procedure ensures that all predicted trajectories respect basic kinematic constraints.

\begin{figure}[h]
    \centering
    \includegraphics[width=\textwidth]{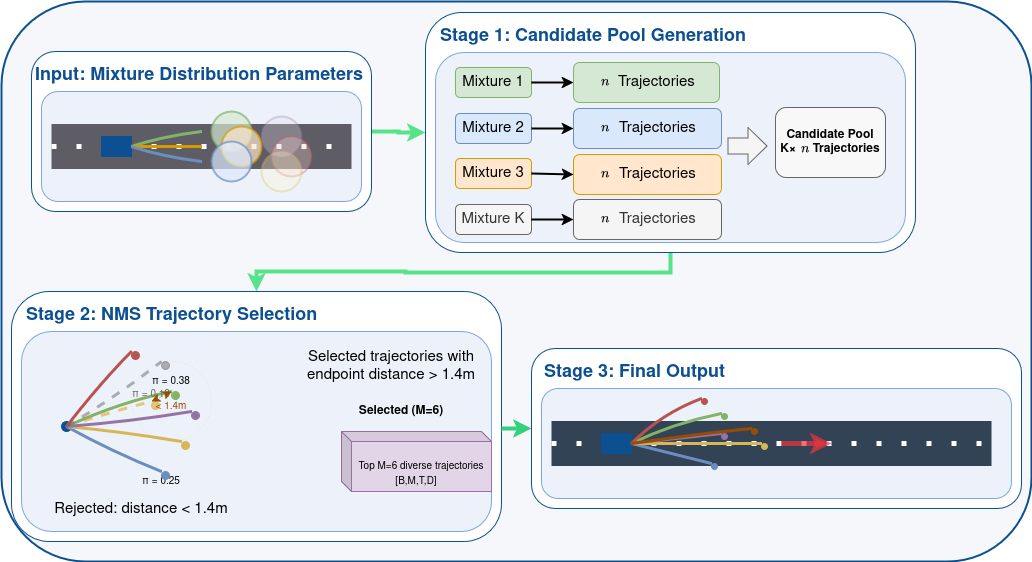}
    \caption{NMS Sampling for Diverse Trajectories. (1) For each mixture component $k \in [1, K]$: (a) Extract $\mu_k \in \mathbb{R}^{T \times D}$ and $\sigma_k \in \mathbb{R}^{T \times D}$; (b) Generate $n$ independent trajectory samples. (2) For each trajectory sample and each timestep $t \in [1, T]$: (i) Create a multivariate normal distribution $\mathcal{N}(\mu_{kt}, \mathrm{diag}(\max(\sigma_{kt}, 10^{-6})))$; (ii) Sample timestep values $x_{kt} \sim \mathcal{N}(\mu_{kt}, \Sigma_{kt})$. (3) Each trajectory inherits probability from its mixture: $p(\mathrm{traj}) = \pi_k$ for trajectories from mixture $k$.}
    \label{fig:NMS}
\end{figure}

\paragraph{Trajectory Sampling with NMS.}
To obtain diverse yet representative trajectories, we employ Non-Maximum Suppression (NMS) sampling (Fig.~\ref{fig:NMS}), implemented in the \texttt{nms\_sampling} method. For each mixture component $k$, we sample multiple candidate trajectories from $\mathcal{N}(\boldsymbol{\mu}_k, \boldsymbol{\Sigma}_k)$. The candidates are then sorted according to their mixture weights $\pi_k$. We greedily select trajectories, ensuring a minimum distance between the endpoints of selected trajectories to promote diversity:
\[
\text{Select trajectory } i \text{ if } \|\mathbf{y}_i(T_{\text{future}}) - \mathbf{y}_j(T_{\text{future}})\|_2 > d_{\min} \quad \forall j \in \text{selected}
\]
where $d_{\min}$ is a minimum distance threshold. 
This approach ensures that the final set of predicted trajectories is both diverse and physically feasible.


\subsection{MPC-Based Trajectory Refinement}
To further ensure dynamical feasibility, we refine the sampled trajectories using a MPC approach~\cite{falcone2007predictive} (Details in Appendix).
We adopt an Ackermann steering vehicle model, where the state vector is defined as $\mathbf{x} = [x, y, \psi, v_x, v_y, \dot{\psi}, a_x, a_y, \ddot{\psi}]^T$ and the control input as $\mathbf{u} = [a_x, a_y, \dot{\psi}]^T$. The discretized vehicle dynamics are given by:

\[
x_{t+1} = x_t + v_{x,t} \cos(\psi_t) \Delta t,\qquad
y_{t+1} = y_t + v_{x,t} \sin(\psi_t) \Delta t,\qquad
\psi_{t+1} = \psi_t + \dot{\psi}_t \Delta t 
\]
\[
v_{x,t+1} = v_{x,t} + a_{x,t} \Delta t,\qquad  \qquad
v_{y,t+1} = v_{y,t} + a_{y,t} \Delta t,\qquad  \qquad
\dot{\psi}_{t+1} = u_{\dot{\psi},t}
\]
\[
a_{x,t+1} = u_{a_x,t},\qquad \qquad \qquad \quad
a_{y,t+1} = u_{a_y,t},\qquad \quad \quad \qquad
\ddot{\psi}_{t+1} = \frac{u_{\dot{\psi},t} - \dot{\psi}_t}{\Delta t}
\]

Given a sampled trajectory, we formulate an MPC optimization problem to find a sequence of control inputs $\{\mathbf{u}_t\}_{t=0}^{T_{\text{future}}-1}$ that minimizes the following cost function:
\[
J(\mathbf{U}) = \sum_{t=0}^{T_{\text{future}}-1} \left[ (\mathbf{x}_{t+1} - \mathbf{x}^{\text{ref}}_{t+1})^T \mathbf{Q} (\mathbf{x}_{t+1} - \mathbf{x}^{\text{ref}}_{t+1}) + \mathbf{u}_t^T \mathbf{R} \mathbf{u}_t \right]
\]
subject to the following constraints:
\RNum{1}. \textbf{Control constraints:} $\mathbf{u}_{\min} \leq \mathbf{u}_t \leq \mathbf{u}_{\max}$.
\RNum{2}. \textbf{Dynamics constraints:} $\mathbf{x}_{t+1} = f(\mathbf{x}_t, \mathbf{u}_t)$
where $\mathbf{x}^{\text{ref}}$ is the reference trajectory, $\mathbf{Q}$ and $\mathbf{R}$ are weighting matrices for state error and control effort, and $f$ denotes the vehicle dynamics model.

The optimization is solved using Sequential Least Squares Programming as implemented in the \texttt{MPC}. This process yields a smooth and dynamically feasible trajectory that can be executed by the vehicle.



\subsection{Loss Function}
Model is trained with a composite loss :
$\mathcal{L} = \mathcal{L}_{pos} + \lambda_{yaw} \mathcal{L}_{yaw} + \lambda_{unc} \mathcal{L}_{unc} + \lambda_{KL} \mathcal{L}_{KL}$
where:
\begin{itemize}
    \item $\mathcal{L}_{pos} = \frac{1}{B \times S} \sum_{b=1}^{B} \sum_{s=1}^{S} \|\hat{\mathbf{p}}_{b,s} - \mathbf{p}^{gt}_{b}\|_1$ is the L1 position error between smoothed trajectories and ground truth, averaged over batch size $B$ and sampled trajectories $S$
    
    \item $\mathcal{L}_{yaw} = \frac{1}{B \times S} \sum_{b=1}^{B} \sum_{s=1}^{S} \|\hat{\boldsymbol{\psi}}_{b,s} - \boldsymbol{\psi}^{gt}_{b}\|_1$ is yaw error between predicted and ground truth
    
    \item $\mathcal{L}_{unc} = -\frac{1}{B \times K \times T \times D} \sum \log(\boldsymbol{\sigma} + 10^{-6})$ is the negative log variance term that encourages appropriate uncertainty estimation while preventing vanishing variances
    
    \item $\mathcal{L}_{KL} = \frac{1}{B} \sum_{b=1}^{B} D_{KL}\Big(\text{Cat}(\boldsymbol{\pi}_b) \big\| \text{Uniform}(K)\Big)$ is the KL divergence between mixture weights and a uniform distribution, promoting balanced mode usage across all $K$ mixtures
\end{itemize}



\section{Experiments}


\subsection{Datasets}
We evaluate our method by predicting trajectories 3 seconds into the future (15 frames at 5 Hz), modeling both positions and headings. The algorithm was trained on five compute nodes, each equipped with two Intel Xeon Cascade Lake processors (20 cores at 2.1 GHz) and four NVIDIA A100 GPUs.

Our evaluation is conducted on two publicly available datasets:

\RNum{1}. \textbf{Lyft}~\cite{houston2020one}: This large-scale dataset comprises over 1,000 hours of autonomous driving data collected across diverse urban environments. We adopt the standard training/validation split, utilizing 25 and 5 seconds of driving data per scenario, respectively.

\RNum{2}. \textbf{nuPlan}~\cite{caesar2021nuplan}: A challenging dataset that captures complex multi-agent interactions over 1,300 km of driving data. It is particularly well-suited for assessing the physical feasibility of predicted trajectories.

\begin{table*}[h]
\caption{Comparison with baselines of closed-loop performance on the Lyft and nuPlan datasets. }

\centering
\resizebox{\textwidth}{!}
{
\begin{NiceTabular}{@{}lcccccc@{}}
\toprule
{\bf Model} & {\bf Dataset} & {\bf  \#Params}  &{\bf Collision (\%)} & {\bf Off-road (\%)} & {\bf Discomfort (\%)} & {\bf L2(m)} 
\\ \midrule
Raster-perturb~\cite{houston2020one, he2016deep}\tabularnote{There is no variance in Raster-perturb, BC-perturb, and UrbanDriver since the previous work evaluated the deterministic pre-trained models in a deterministic simulator.} & Lyft & 23.6M & 15.48 & 5.06 & \textbf{4.00}  &  5.90\\
BC-perturb~\cite{scheel2021urban}\tabularnote{There is no variance in Raster-perturb, BC-perturb, and UrbanDriver since the previous work evaluated the deterministic pre-trained models in a deterministic simulator.} & Lyft & 1.8M & 9.38 &  6.77 & 39.10 &  4.77 \\
UrbanDriver~\cite{scheel2021urban}\tabularnote{There is no variance in Raster-perturb, BC-perturb, and UrbanDriver since the previous work evaluated the deterministic pre-trained models in a deterministic simulator.} & Lyft & 1.8M &  13.28 &  7.27 & 39.41 &  5.74 \\
\midrule
TD3+BC~\cite{fujimoto2021minimalist}  & Lyft & 2.8M & 22.53\,$\pm$\,1.76  & 15.21\,$\pm$\,0.97   & 4.86\,$\pm$\,0.47 & 6.34\,$\pm$\,0.41   \\
Vector-Chauffeur~\cite{bansal2019chauffeur} & Lyft & \textbf{1.5M} &  10.12\,$\pm$\,0.23 & 3.40\,$\pm$\,0.32 & 5.42\,$\pm$\,0.44 & 5.03\,$\pm$\,0.43     \\
CCIL~\cite{guo2023ccilcontextconditionedimitationlearning} & Lyft & \textbf{1.5M} & 6.46\,$\pm$\,0.43 & 5.54\,$\pm$\,0.21 & 6.54\,$\pm$\,0.27  & 5.23\,$\pm$\,0.47    \\
\midrule
\textbf{PhysVarMix(ours)} & Lyft & 2.5M  & \textbf{3,14\,$\pm$\,0.31}  & \textbf{0.87\,$\pm$\,0.27}  & \textbf{3.43\,$\pm$\,0.33} & \textbf{1.56\,$\pm$\,0.36}  \\
\toprule
\midrule
LaneGCN-perturb~\cite{liang2020learning} & nuPlan & 2.0M & 60.63\,$\pm$\,2.34  & 34.25\,$\pm$\,1.65 & 17.26\,$\pm$\,1.80 & 21.21\,$\pm$\,1.81   \\
TD3+BC~\cite{fujimoto2021minimalist} & nuPlan & 2.8M & 39.12\,$\pm$\,2.21  & 18.59\,$\pm$\,1.04 & 10.56\,$\pm$\,0.95 &  15.04\,$\pm$\,1.62  \\
Vector-Chauffeur~\cite{bansal2019chauffeur} & nuPlan & \textbf{1.5M} & 24.12\,$\pm$\,1.37 & 10.11\,$\pm$\,0.62 & 12.53\,$\pm$\,1.17 & 6.12\,$\pm$\,0.87  \\
CCIL~\cite{guo2023ccilcontextconditionedimitationlearning}\tabularnote{The CCIL code~\cite{guo2023ccilcontextconditionedimitationlearning} is incomplete, lacking the training component. We implemented this part based on the paper's description. The completed version of the CCIL code and our own implementation will be released.}& nuPlan  & \textbf{1.5M} &9.34\,$\pm$\,1.32 & 5.58\,$\pm$\,0.41 & 3.72\,$\pm$\,0.45 & 4.63\,$\pm$\,0.41 \\
\midrule
\textbf{PhysVarMix (ours)}& nuPlan  & 2.5M  & \textbf{6.83\,$\pm$\,0.74}  & \textbf{1.26\,$\pm$\,0.33}  & \textbf{1.18\,$\pm$\,0.58} & \textbf{2.25\,$\pm$\,0.24}  \\
\midrule
\bottomrule
\end{NiceTabular}
}
\label{tab:result}
\end{table*}

\begin{table*}[h]
\caption{Ablation study on closed-loop performance on the Lyft dataset. Each value is reported as mean $\pm$ standard deviation over 3 runs. Lower is better for all metrics. The "Data.Std" refers to the standard deviation used for Gaussian noise augmentation of the agent's centroid (position).}
\centering
\resizebox{\textwidth}{!}
{
\begin{tabular}{@{}lcc|cccc@{}}
\toprule
\multicolumn{1}{@{}l}{\bf Model} & \multicolumn{1}{l}{\bf Data.Std} & \multicolumn{1}{l}{\bf Ego}  &\multicolumn{1}{l}{\bf Collision (\%)} & \multicolumn{1}{l}{\bf Off-road (\%)} & \multicolumn{1}{l}{\bf Discomfort (\%)} & \multicolumn{1}{l@{}}{\bf L2 (m)}
\\
\midrule
Explicit ego features        & --   & \checkmark & 14.15\,$\pm$\,1.02 & 14.30\,$\pm$\,3.10 & 0.68\,$\pm$\,0.18 & 5.97\,$\pm$\,0.62 \\
Ego dropout                 & --   & \checkmark & 11.80\,$\pm$\,1.65 & 6.21\,$\pm$\,1.01  & 0.74\,$\pm$\,0.23 & 4.61\,$\pm$\,0.53  \\
Ego coordinate only         & --   & \checkmark & 12.45\,$\pm$\,1.57 & 11.10\,$\pm$\,1.48  & 1.08\,$\pm$\,0.07 & 4.19\,$\pm$\,0.13 \\
\midrule
No augmentation (std=0)     & 0    & --         & 8.12\,$\pm$\,0.41  & 3.25\,$\pm$\,0.29  & 1.03\,$\pm$\,0.12 & 3.89\,$\pm$\,0.38 \\
Augmentation std=1          & 1    & --         & 3.89\,$\pm$\,0.22  & 1.19\,$\pm$\,0.19  & 2.21\,$\pm$\,0.18 & 2.37\,$\pm$\,0.07 \\
Augmentation std=2 \textbf{(Ours)}   & 2    & --         & 3.75\,$\pm$\,0.19 & 0.73\,$\pm$\,0.15 & 4.82\,$\pm$\,0.25 & 1.38\,$\pm$\,0.09 \\
Augmentation std=3          & 3    & --         & 3.91\,$\pm$\,0.15  & 0.59\,$\pm$\,0.12 & 8.12\,$\pm$\,0.29 & 1.03\,$\pm$\,0.05 \\
\midrule
w/o Causal       & 2    & --         & 4.81\,$\pm$\,0.29  & 1.62\,$\pm$\,0.28  & 7.21\,$\pm$\,0.31 & 1.84\,$\pm$\,0.12 \\
w/o MPC Smoothing     & 2    & --         & 4.29\,$\pm$\,0.17  & 2.31\,$\pm$\,0.14  & 93.12\,$\pm$\,0.41 & 1.15\,$\pm$\,0.03 \\
w/o Sector Constraints      & 2    & --         & 3.62\,$\pm$\,0.18  & 2.41\,$\pm$\,0.21  & 4.91\,$\pm$\,0.27 & 1.41\,$\pm$\,0.10 \\
w/o NMS Sampling            & 2    & --         & 4.21\,$\pm$\,0.23  & 0.81\,$\pm$\,0.13  & 2.09\,$\pm$\,0.19 & 2.21\,$\pm$\,0.08 \\
\midrule
w/o Regularization          & 2    & --         & 4.56\,$\pm$\,0.19  & 1.19\,$\pm$\,0.16  & 5.41\,$\pm$\,0.33 & 1.37\,$\pm$\,0.10  \\
w/o Auxiliary Losses        & 2    & --         & 5.02\,$\pm$\,0.32  & 1.17\,$\pm$\,0.07  & 3.67\,$\pm$\,0.34 & 2.13\,$\pm$\,0.09  \\
\bottomrule
\end{tabular}
}
\label{tab:ablation}
\end{table*}

\subsection{Evaluation Metrics}

We evaluate our model using following trajectory prediction metrics:
\begin{itemize}
    \item \textbf{Collision Rate (\%)}: Percentage of scenarios where the SDV's bounding box intersects with any other agent, indicating a safety-critical failure.
    
    \item \textbf{Off-road Rate (\%)}: For nuPlan, we adopt the official metric flagging a scenario when any corner of the SDV exceeds 0.3m from the drivable area. For Lyft, following UrbanDriver, we consider a scenario off-road when lateral deviation from ground truth exceeds 2m.
    
    \item \textbf{Discomfort (\%)}: Percentage of timesteps where absolute acceleration exceeds 3 m/s$^2$, quantifying ride comfort and trajectory smoothness.
    
    \item \textbf{L2 Error (m)}: Average Euclidean distance between predicted and ground truth positions, measuring prediction accuracy.
\end{itemize}

\subsection{Quantitative Evaluation}

Table~\ref{tab:result} presents the main results on the Lyft and nuPlan datasets. Our PhysVarMix model exhibits outstanding results, leading to more accurate environmental understanding and safer trajectory planning on both the challenging Lyft and nuPlan benchmarks.


\subsection{Ablation Studies}

We conduct ablation studies to evaluate the contribution of each component (Table~\ref{tab:ablation}).
Observations:
\begin{enumerate}
    \item Removing sector constraints slightly improves results but significantly reduces physical feasibility, making the predictions less useful for downstream planning.
    \item NMS sampling is critical for introducing diversity into the selection of driving trajectories, ultimately enhancing the off-road navigation rate by prioritizing distinct and viable options.
    \item MPC smoothing contributes to the Discomfort rate by enforcing realistic vehicle dynamics.
    \item The causal structure is non-negligible for predictions, particularly in interactive scenarios.
\end{enumerate}

\subsection{Qualitative Analysis}

Figure~\ref{fig:allvis} presents qualitative examples of our model's predictions in challenging scenarios. Our PhysVarMix framework generates multiple diverse, physically plausible trajectories that accurately capture different possible behaviors while respecting road geometry and agent dynamics.


\subsection{Discussion}

Our experiments demonstrate that PhysVarMix successfully balances prediction accuracy, trajectory diversity, and physical feasibility. The integration of causal network with physics constraints and diverse sampling strategies effectively addresses key limitations in existing approaches.
The ablation studies highlight the complementary nature of our components: the causal transformer provides temporal consistency, NMS sampling ensures diversity, sector constraints maintain physical plausibility, and MPC smoothing refines trajectories for realism.
A particularly interesting finding is that while removing sector constraints can slightly improve raw accuracy metrics, the resulting physically implausible trajectories would be unsuitable for downstream planning modules. This demonstrates the importance of incorporating domain knowledge into learning-based trajectory prediction systems.

\section{Conclusion}

In this work, we proposed PhysVarMix, a novel framework that addresses the fundamental tension between diversity and feasibility in trajectory prediction for autonomous driving. By integrating a causal transformer architecture with physically-informed constraints, our approach successfully generates diverse, temporally consistent, and feasible trajectories across various complex driving scenarios.
Our main contributions are threefold. First, we developed a causal network that effectively captures temporal dependencies in agent-environment interactions while maintaining strict causality in the prediction process. Second, we introduced variational mixture density networks with a novel NMS sampling strategy that enhances trajectory diversity while avoiding redundancy. Third, we implemented explicit physical constraints through sector boundaries and MPC-based smoothing that guarantee the feasibility of predicted trajectories.
The experimental results demonstrate that our approach advances the outstanding performance by bridging the gap between data-driven prediction and physics-based modeling. The model produces diverse, interpretable, and physically consistent trajectories that can be directly integrated into downstream planning modules. The probabilistic nature of our predictions enables autonomous vehicles to reason about uncertainty, leading to safer and more efficient decision-making in complex driving scenarios.




\clearpage

\appendix

\paragraph{Appendix}

\begin{figure}[h]
    \centering
    \includegraphics[width=\textwidth]{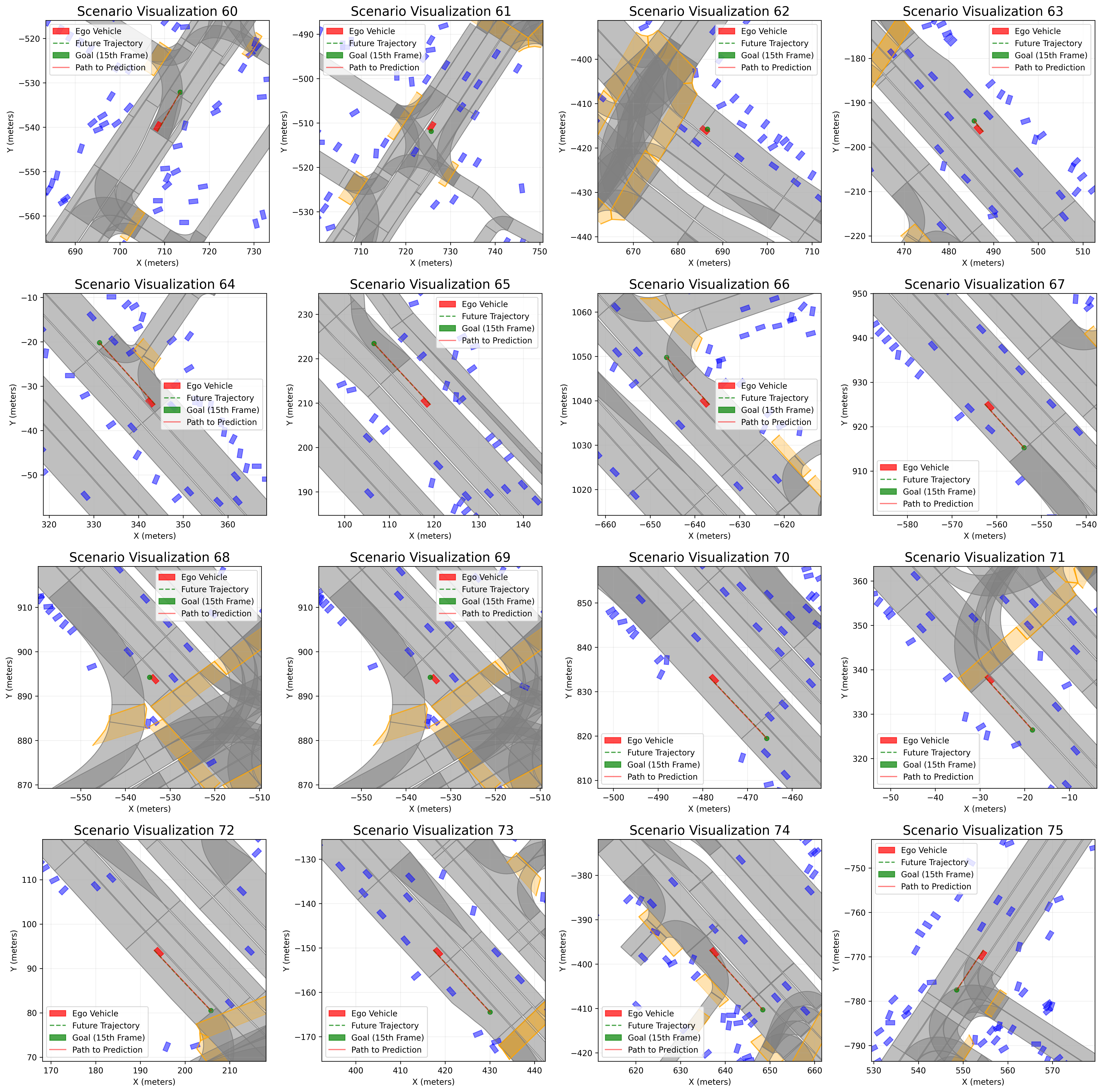}
    \caption{Our PhysVarMix framework is demonstrated through qualitative examples in challenging scenarios, showcasing its ability to generate multiple diverse and physically plausible trajectories. These generated trajectories effectively capture a range of potential agent behaviors while adhering to road geometry and agent dynamics constraints.}
    \label{fig:allvis}
\end{figure}

\section{MPC Controller}

We modeled the vehicle's motion using a kinematic model, focusing on geometric aspects of movement while incorporating essential physical constraints. This approach strikes a balance between fidelity and computational efficiency, particularly suitable for our algorithm.

We represent the vehicle's state using a vector $\mathbf{x} = (x, y, \psi)^T$, where:

    $(x, y)$ denotes the position of the vehicle's center.

    $\psi$ represents the vehicle's body angle, defined as the angle between its longitudinal axis (passing through the wheel centers) and the x-axis.

The control inputs to the vehicle are represented by the vector $\mathbf{u} = (v, \delta)^T$, where:

    $v$ is the vehicle's velocity, assumed to be equivalent to the wheel velocity in the absence of slip.

    $\delta$ denotes the steering angle of the front wheels.

\begin{figure}[h]
    \begin{minipage}[t]{0.5\linewidth}
        \centering
        \includegraphics[width=\textwidth]{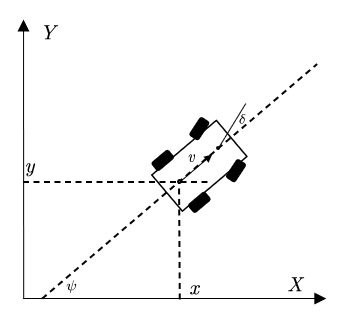}
        \centerline{\footnotesize(a) Vehicle Modeling}
    \end{minipage}%
    \begin{minipage}[t]{0.5\linewidth}
        \centering
        \includegraphics[width=\textwidth]{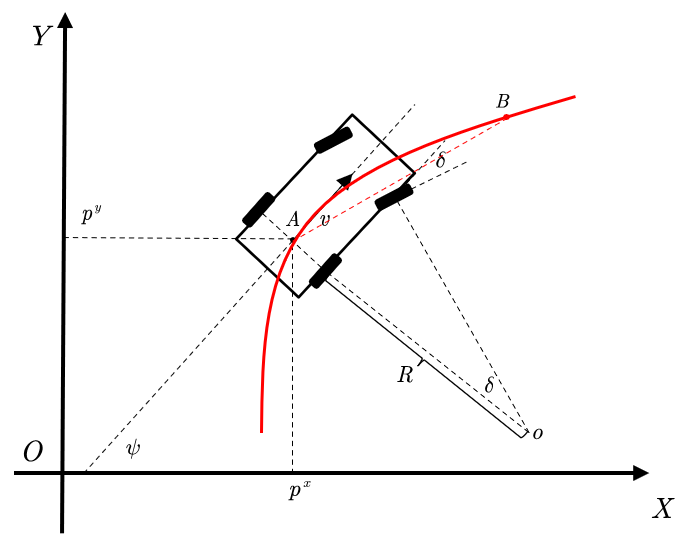}
        \centerline{\footnotesize(b) Trajectory Tracking}
    \end{minipage}
    \caption{System Model of the trajectory tracking problem of an autonomous vehicle.}
    \label{FigControl}
\end{figure}

Fig.~\ref{FigControl}(b) illustrates the trajectory tracking scenario, where $R$ represents the vehicle's turning radius, a crucial parameter determined by the steering angle and vehicle dynamics.

Based on these definitions and geometric relationships, we derive the kinematic model governing the vehicle's motion. This model, detailed in the subsequent section, serves as the foundation for our MPC controller, enabling accurate prediction of the vehicle's future states and the generation of optimal control inputs for trajectory following.

\begin{equation}
    \left[ \begin{array}{c}
            \dot{x}    \\
            \dot{y}    \\
            \dot{\psi} \\
        \end{array} \right] =\left[ \begin{array}{c}
            \cos \psi      \\
            \sin \psi      \\
            \tan \delta /l \\
        \end{array} \right] v
\end{equation}
Obviously, it is a nonlinear system, and there is a coupling relationship between the inputs $v$ and $\delta$. In order to simplify the controller design, linearization and discretization are taken into account.
Equation (1) can be regarded as a first-order continuous differential equation:
\begin{equation}
    \dot{\mathbf{x}}=f\left( \mathbf{x} ,\mathbf{u} \right)
\end{equation}
The reference trajectory is $(\mathbf{x}_r, \mathbf{u}_r)$ and for the system there is $\dot{\mathbf{x}}_r=f\left( \mathbf{x}_r ,\mathbf{u}_r \right)$. The system can be approximated in a Taylor series at $(\mathbf{x}_r ,\mathbf{u}_r)$:
\begin{equation}
    \dot{\mathbf{x}}\approx f\left( \mathbf{x}_r ,\mathbf{u}_r \right) +\frac{\partial f}{\partial \mathbf{x}}\left(  \mathbf{x} -\mathbf{x}_r  \right) +\frac{\partial f}{\partial \mathbf{u}}\left( \mathbf{u}-\mathbf{u}_r \right)
\end{equation}
After linearization, the system can be written as follows:
\begin{equation}
    \begin{array}{c}
        \dot{\mathbf{x}}=A\mathbf{x} +B\mathbf{u}                                                                                                     \\
        A=\left[ \begin{matrix}
                         0 & 0 & -v\sin \!\:\psi \\
                         0 & 0 & v\cos \!\:\psi  \\
                         0 & 0 & 0               \\
                     \end{matrix} \right] B=\left[ \begin{matrix}
                                                       \cos \!\:\psi              & 0                                                     \\
                                                       \sin \!\:\psi              & 0                                                     \\
                                                       \frac{\tan \!\:\delta }{l} & \frac{v}{\mathrm{l}\cos ^2\!\:\left( \delta  \right)} \\
                                                   \end{matrix} \right] \\
    \end{array}
\end{equation}
The goal is to control the vehicle to complete the trajectory tracking; in other words, control the system states converge to reference trajectory. Define $\mathrm{\tilde{\mathbf{x}}}=\mathrm{\mathbf{x}}-\mathrm{\mathbf{x}}_r$ and the discretization can be done according to the following equations in which $T$ is the length of the time steps:
\begin{equation}
    \left\{ \begin{array}{l}
        \tilde{A}=I_{n\times n}+TA \\
        \tilde{B}=TB               \\
    \end{array} \right.
\end{equation}
Then, the discrete form of the system is written as:
\begin{equation}
    \begin{array}{c}
        \tilde{\mathbf{x}}_{k+1}=\tilde{A}\tilde{\mathbf{x}}_k+\tilde{B}\tilde{\mathbf{u}}_k                  \\
        \mathrm{\tilde{\mathbf{x}}}=\left[ \begin{array}{c}
                                                   x-x_r         \\
                                                   y-y_r         \\
                                                   \psi -\psi _r \\
                                               \end{array} \right] \tilde{A}=\left[ \begin{matrix}
                                                                                        1 & 0 & -v_r\sin \!\:\psi _rT \\
                                                                                        0 & 1 & v_r\cos \!\:\psi _rT  \\
                                                                                        0 & 0 & 1                     \\
                                                                                    \end{matrix} \right] \\
        \tilde{B}=\left[ \begin{matrix}
                                 \cos \!\:\!\:\psi _rT         & 0                                                          \\
                                 \sin \!\:\!\:\psi _rT         & 0                                                          \\
                                 \frac{\tan \!\:\delta _rT}{l} & \frac{v_rT}{\mathrm{l}\cos ^2\!\:\left( \delta _r \right)} \\
                             \end{matrix} \right]
        \tilde{\mathbf{u}}_k=\left[\begin{array}{c}
                                           v-v_r \\
                                           \delta - \delta_r
                                       \end{array}
            \right]
    \end{array}
\end{equation}

\section{Data Visualization and Processing}

In our autonomous driving research, we implement comprehensive visualization and processing pipelines for understanding the spatial relationships between different elements in driving scenes. Our framework processes the Lyft dataset which contains rich map information and agent dynamics.

\paragraph{Map Visualization}

The Lyft dataset contains two primary map elements: lanes and crosswalks. Our visualization pipeline renders these elements in a bird's-eye view representation centered on the ego vehicle. Lanes are visualized as filled polygons constructed by connecting the left and right lane boundaries, with gray coloring and semi-transparency to distinguish different lanes while maintaining visual clarity. Crosswalks are represented as orange polygons with higher opacity to highlight areas of potential pedestrian interactions.

For each lane element, we extract coordinate information using the Lyft map API, obtaining both left and right boundary points. These boundaries are then rendered to create a complete representation of the drivable area. Similarly, crosswalk polygons are extracted and visualized with distinct coloring to emphasize these critical map elements.

\paragraph{Agent Representation}

Agents in the scene are rendered as rectangles with proper orientation based on their yaw angle. The ego vehicle is prominently displayed in red with higher opacity to distinguish it from other agents, which are rendered in blue with lower opacity. This visual hierarchy helps focus attention on the primary vehicle while maintaining awareness of surrounding traffic participants.

\paragraph{Trajectory Visualization}

A key aspect of our visualization is the representation of future trajectories. We extract sequential ego positions from future frames (up to 15 frames ahead) and render them as dashed green lines to indicate the ground truth path. The final position is highlighted with a green circular marker to denote the goal position. Additionally, we connect the current ego position to the prediction position with a red line to visualize the direct path.

\paragraph{Processing Pipeline}

Our data processing pipeline efficiently extracts relevant information from the Zarr dataset structure used by Lyft. For each visualization:

\begin{itemize}
    \item We retrieve scene information and corresponding frames based on the scene index
    \item We extract ego vehicle position and orientation from the frame data
    \item We process map elements within a relevant region around the ego vehicle
    \item We extract other agent information from the same frame
    \item We compute future trajectories by analyzing subsequent frames in the scene    
\end{itemize}

The visualization is spatially bounded to focus on a 100m × 100m area centered on the ego vehicle, providing sufficient context of the surrounding environment while maintaining visual clarity. This bounded view creates a consistent representation across different scenes, facilitating comparative analysis.

Our visualization approach supports both individual scene analysis and batch processing of multiple scenes, enabling systematic evaluation of model performance across diverse driving scenarios. The resulting visualizations provide intuitive insights into the spatial relationships between map elements, dynamic agents, and predicted trajectories.

\section{Model Parameters}

For both datasets, the same model architecture is used, whose hyper-parameters are listed in~\ref{tab:Hyperparameters}.

\begin{table}[t]
\caption{Hyper-parameters for the Model}
\label{tab:Hyperparameters}
\begin{center}
\begin{tabular}{lc}
\multicolumn{1}{c}{\bf Hyper-parameter} & \multicolumn{1}{c}{\bf Value} \\
\hline \\
\multicolumn{2}{l}{\textbf{General Transformer Parameters}} \\
    Future steps $T$ (prediction horizon) &  15 \\
    All Transformers dropout &  0.1 \\
    All Transformers head number & 8 \\
    All Transformers hidden size ($d_{model}$) &  128 \\
\multicolumn{2}{l}{\textbf{Scene Encoder Parameters}} \\
    Local Transformer layer number & 6\\
    Global Transformer layer number  &  6 \\
    Lane feature dimension (lane\_feat\_dim) & 10 \\
    Crosswalk feature dimension (cross\_feat\_dim) & 5 \\
    Agent feature dimension (agent\_feat\_dim) & 7 \\
    History frames for agents & 12 \\
\multicolumn{2}{l}{\textbf{Causal Network Parameters}} \\
    Causal Network layer number & 6\\
    Causal Network history length $H$ (causal\_len) &  15 \\
    Causal Network interval $I$&  2 \\
\multicolumn{2}{l}{\textbf{Variational Bayes Mixture Model Parameters}} \\
    Number of mixtures $K$ & 6 \\
    Latent dimension $v$ & 16 \\ 
    Auxiliary weight $\mu$ & 0.3 \\ 
\multicolumn{2}{l}{\textbf{Loss Function and Regularization}} \\
    Regularization weight $\lambda$ (optimizer weight\_decay) & 0.001 \\
    KL divergence loss weight (Categorical) & 0.01 \\
    Uncertainty loss weight & 0.1 \\
\multicolumn{2}{l}{\textbf{MPC Smoother Weights}} \\
    MPC State: Positional acceleration weight ($\text{acc\_w}, \eta_{va}$) & 0.1\\
    MPC State: Angular velocity weight ($\text{yaw\_vel\_w}, \eta_{\omega}$) &  0.1 \\
    MPC State: Angular acceleration weight ($\text{yaw\_acc\_w}, \eta_{\alpha}$) &  0.1 \\
    MPC State: Yaw weight ($\text{yaw\_w}$) & 1 \\
    MPC Control: Effort weight ($\text{control\_w}$) & 0.01 \\
    MPC Control: Yaw rate weight ($\text{yaw\_control\_w}$) & 0.001 \\
    MPC positional jerk weight $\eta\_{vj}$ & 0.01 \\ 
\multicolumn{2}{l}{\textbf{MPC Solver Parameters}} \\
    MPC Time step $dt$ (step\_time) & 0.1 \\
    MPC Wheelbase & 3.089  \\
\multicolumn{2}{l}{\textbf{Model Architecture \& Input/Output Dimensions}} \\
    Action dimension (act\_dim) & 3 \\
\multicolumn{2}{l}{\textbf{NMS Sampling Parameters}} \\
    Number of samples (num\_samples) & 6 \\
    NMS distance threshold & 1.4 \\
\multicolumn{2}{l}{\textbf{Other Model Design Choices}} \\
    Sector boundary max angle deviation & $\pi/4$ rad  \\
    Sector boundary max radius & 10.0 m \\
\end{tabular}
\end{center}
\end{table}

\section{Mathematical Definitions}

This section provides a detailed step-by-step mathematical derivation of the trajectory prediction model's components.

\subsection{Scene Encoding}
The goal is to transform raw inputs $f_{raw}$ into a scene embedding $\bm{s}_t \in \mathbb{R}^{d_{model}}$.

\subsubsection{Individual Feature Embedding}
Let $\bm{x}$ be a generic input feature vector for an entity (ego, agent, or map point).
\begin{itemize}[leftmargin=*]
    \item \textbf{Multi-Layer Perceptron (MLP)}: Used for ego and agent features.
    An MLP, $\text{MLP}(\bm{x})$, typically comprises:
    \begin{enumerate}[label=(\roman*), leftmargin=2em]
        \item First linear transformation: $\bm{z}_1 = \bm{W}_1 \bm{x} + \bm{b}_1$.
        \item Layer Normalization: $\bm{z}_{1,norm} = \text{LayerNorm}(\bm{z}_1)$.
        \item Activation: $\bm{a}_1 = \text{ReLU}(\bm{z}_{1,norm})$.
        \item Second linear transformation (output embedding): $\bm{h} = \bm{W}_2 \bm{a}_1 + \bm{b}_2 \in \mathbb{R}^{d_{model}}$.
    \end{enumerate}
    Thus, $\bm{h}_{ego} = \text{MLP}_{ego}(f_{ego})$ and $\bm{h}_{agent_i} = \text{MLP}_{agent}(f_{agent_i})$.

    \item \textbf{LinearWithNorm}: Used for map polyline points $\bm{p}_m$.
    \[ \bm{p}'_{m} = \text{ReLU}(\text{LayerNorm}(\bm{W}_{map} \bm{p}_m + \bm{b}_{map})) \in \mathbb{R}^{d_{model}} \]
\end{itemize}

\subsubsection{Map Element Processing (\texttt{MapTransformer})}
For a map polyline (e.g., lane) represented by a sequence of embedded points $\bm{P}' = [\bm{p}'_1, ..., \bm{p}'_L]$, where each $\bm{p}'_m \in \mathbb{R}^{d_{model}}$.
This sequence is processed by a local Transformer Encoder, $\text{TransformerEncoder}_{local}(\bm{P}')$. A Transformer Encoder layer consists of:
\begin{enumerate}[label=(\alph*), leftmargin=2em]
    \item \textbf{Multi-Head Self-Attention (MHSA)}:
    For each head $j \in \{1, ..., N_{heads}\}$ (where $N_{heads}=8$):
    \begin{itemize}[leftmargin=1.5em]
        \item Linearly project $\bm{P}'$ to Query ($\bm{Q}_j$), Key ($\bm{K}_j$), Value ($\bm{V}_j$) matrices:
        $\bm{Q}_j = \bm{P}' \bm{W}_{Q,j}$, $\bm{K}_j = \bm{P}' \bm{W}_{K,j}$, $\bm{V}_j = \bm{P}' \bm{W}_{V,j}$.
        Each $\bm{W}$ is a learnable weight matrix. The dimension of queries, keys, and values per head is $d_k = d_{model}/N_{heads}$.
        \item Scaled Dot-Product Attention:
        \[ \text{Attention}_j(\bm{Q}_j, \bm{K}_j, \bm{V}_j) = \text{softmax}\left(\frac{\bm{Q}_j \bm{K}_j^T}{\sqrt{d_k}} + \bm{M}_{pad}\right) \bm{V}_j = \bm{H}_j \]
        $\bm{M}_{pad}$ is a padding mask to ignore invalid points in the polyline.
    \end{itemize}
    \item Concatenate heads and project: $\bm{H}_{MHSA} = \text{Concat}(\bm{H}_1, ..., \bm{H}_{N_{heads}}) \bm{W}_O$.
    \item Add \& Norm (Residual Connection and Layer Normalization):
    $\bm{A} = \text{LayerNorm}(\bm{P}' + \bm{H}_{MHSA})$.
    \item \textbf{Feed-Forward Network (FFN)}: Applied position-wise.
    $\text{FFN}(\bm{a}) = \text{ReLU}(\bm{a} \bm{W}_{f1} + \bm{b}_{f1}) \bm{W}_{f2} + \bm{b}_{f2}$.
    $\bm{F} = \text{FFN}(\bm{A})$. (Applied row-wise to $\bm{A}$)
    \item Add \& Norm: $\bm{P}_{encoded\_layer} = \text{LayerNorm}(\bm{A} + \bm{F})$.
\end{enumerate}
This stack of operations (a-d) is repeated for \texttt{local\_num\_layers} (6). Let the final output be $\bm{P}_{encoded} \in \mathbb{R}^{L \times d_{model}}$.
The polyline feature is then $\bm{h}_{poly} = \max_{m}(\bm{P}_{encoded}[m,:]) \in \mathbb{R}^{d_{model}}$.

\subsubsection{Global Context Aggregation}
The set of all entity embeddings $\bm{H}_{all} = [\bm{h}_{ego}, \bm{h}_{agent_1}, ..., \bm{h}_{poly_1}, ...]$ is processed by a global Transformer Encoder, $\text{TransformerEncoder}_{global}(\bm{H}_{all})$, with \texttt{global\_num\_layers} (6). Its structure is identical to the local Transformer Encoder described above (MHSA, Add\&Norm, FFN, Add\&Norm), operating on the sequence of entity embeddings.
The scene embedding is $\bm{s}_t = (\text{TransformerEncoder}_{global}(\bm{H}_{all}))[0] \in \mathbb{R}^{d_{model}}$.

\subsection{Causal Temporal Encoding}
Input: Sequence of scene embeddings $\bm{S}_{hist} = [\bm{s}_{t-H'+1}, ..., \bm{s}_t]$.
\begin{enumerate}[label=(\alph*), leftmargin=2em]
    \item Positional Encoding: $\bm{e}_i = \bm{s}_i + \text{Embedding}_{time}(i)$. Let $\bm{E}_{seq} = [\bm{e}_1, ..., \bm{e}_{H'}]$.
    \item Causal Network Encoder: $\bm{O}_{transformer} = \text{TransformerEncoder}_{causal}(\bm{E}_{seq})$.
    This encoder is similar to the one described in D1.2(a), but the Scaled Dot-Product Attention in MHSA uses a causal mask $\bm{M}_{causal}$ in addition to any padding mask:
    \[ \text{Attention}_j(\bm{Q}_j, \bm{K}_j, \bm{V}_j) = \text{softmax}\left(\frac{\bm{Q}_j \bm{K}_j^T}{\sqrt{d_k}} + \bm{M}_{causal} + \bm{M}_{pad}\right) \bm{V}_j \]
    $\bm{M}_{causal}$ ensures that position $i$ cannot attend to positions $j > i$.
    \item Output Context Vector $\bm{o}_t$:
    \[ \bm{o}_t = \begin{cases} \frac{1}{H'} \sum_{i=1}^{H'} \bm{O}_{transformer}[i,:] & \text{if training} \\ \bm{O}_{transformer}[H',:] & \text{if inference} \end{cases} \]
\end{enumerate}

\subsection{Probabilistic Trajectory Prediction (Variational Bayes Mixture Model)}
Predicts $K$ GMM components over $T_{fut}$ steps.
\subsubsection{GMM Parameters from Context}
\begin{itemize}[leftmargin=*]
    \item Means $\bm{\mu}_k \in \mathbb{R}^{T_{fut} \times d_{act}}$:
    The vector of all means $\text{vec}(\bm{M}_{\mu}) \in \mathbb{R}^{K \cdot T_{fut} \cdot d_{act}}$ is obtained by a linear transformation:
    \[ \text{vec}(\bm{M}_{\mu}) = \bm{W}_{\mu} \bm{o}_t + \bm{b}_{\mu} \]
    This is then reshaped to $[\bm{\mu}_1, ..., \bm{\mu}_K]$.
    \item Log-Variances $\log \bm{\sigma}_k^2 \in \mathbb{R}^{T_{fut} \times d_{act}}$:
    Similarly, $\text{vec}(\log \bm{M}_{\sigma^2}) = \bm{W}_{\sigma} \bm{o}_t + \bm{b}_{\sigma}$.
    Then, $\bm{\sigma}_k^2[t,d] = \exp(\text{clamp}(\log \bm{\sigma}_k^2[t,d], \max=10))$.
\end{itemize}

\subsubsection{Latent Variable}
\begin{itemize}[leftmargin=*]
    \item Approximate Posterior $q(\bm{v}|\bm{o}_t) = \mathcal{N}(\bm{v} | \bm{\mu}_v, \text{diag}(\bm{\sigma}_v^2))$:
    \begin{itemize}[leftmargin=1.5em]
        \item $\bm{h}_{\mu_v}, \bm{c}_{\mu_v} = \text{LSTMCell}_{\mu_v}(\bm{o}_t, (\bm{h}_{init}, \bm{c}_{init}))$. Then $\bm{\mu}_v = \bm{h}_{\mu_v}$. (Assuming $\bm{o}_t$ is input to LSTM as a single step).
        \item $\log \bm{\sigma}_v^2 = \bm{W}_{v\sigma} \bm{o}_t + \bm{b}_{v\sigma}$. Then $\bm{\sigma}_v^2 = \exp(\text{clamp}(\log \bm{\sigma}_v^2, \max=10))$.
    \end{itemize}
    \item Sampling: $\bm{v}_{sample} = \bm{\mu}_v + \bm{\sigma}_v \odot \bm{\epsilon}$, where $\bm{\epsilon} \sim \mathcal{N}(0, I)$ (training). $\bm{v}_{sample} = \bm{\mu}_v$ (inference).
    \item Temporal Propagation of $\bm{v}_{sample}$:
    Input $\bm{v}_{sample}$ to $\text{LSTM}_{v\_trans}$ for $T_{fut}$ steps (e.g., by repeating $\bm{v}_{sample}$ as input or using it as initial hidden state and zero inputs).
    For $j=1, ..., T_{fut}$: $\bm{h}_{v,j}, \bm{c}_{v,j} = \text{LSTMCell}_{v\_trans}(\bm{x}_{in,j}, (\bm{h}_{v,j-1}, \bm{c}_{v,j-1}))$.
    Output sequence $\bm{v}_{seq} = [\bm{h}_{v,1}, ..., \bm{h}_{v,T_{fut}}]$.
\end{itemize}

\subsubsection{Mixture Probabilities}
\[ \bm{z}_{\pi} = \text{MLP}_{\pi}(\bm{o}_t) \]
\[ \pi_k = \frac{\exp(z_{\pi,k})}{\sum_{j=1}^K \exp(z_{\pi,j})} \quad \text{for } k=1,...,K \]

\subsection{Trajectory Sampling and Refinement}
\subsubsection{Sector Boundary Application}
For an ego pose $(\bm{p}_{ego}, \psi_{ego})$ and a predicted point $\bm{p}_{pred}=(x_{pred}, y_{pred})$:
\begin{enumerate}[label=(\alph*), leftmargin=2em]
    \item Relative position in ego frame:
    $\Delta x_{ego} = (x_{pred} - x_{ego})\cos\psi_{ego} + (y_{pred} - y_{ego})\sin\psi_{ego}$
    $\Delta y_{ego} = -(x_{pred} - x_{ego})\sin\psi_{ego} + (y_{pred} - y_{ego})\cos\psi_{ego}$
    \item Convert to polar in ego frame: $d = \sqrt{\Delta x_{ego}^2 + \Delta y_{ego}^2}$, $\alpha_{rel} = \text{atan2}(\Delta y_{ego}, \Delta x_{ego})$.
    \item Clamp: $d' = \min(d, R_{max})$, $\alpha'_{rel} = \text{clamp}(\alpha_{rel}, -\Delta\psi_{max}, \Delta\psi_{max})$.
    \item Convert back to Cartesian in ego frame: $\Delta x'_{ego} = d' \cos(\alpha'_{rel})$, $\Delta y'_{ego} = d' \sin(\alpha'_{rel})$.
    \item Convert back to world frame:
    $x'_{pred} = x_{ego} + \Delta x'_{ego} \cos\psi_{ego} - \Delta y'_{ego} \sin\psi_{ego}$
    $y'_{pred} = y_{ego} + \Delta x'_{ego} \sin\psi_{ego} + \Delta y'_{ego} \cos\psi_{ego}$
\end{enumerate}

\subsubsection{Non-Maximum Suppression (NMS) Sampling}
\begin{enumerate}[label=(\alph*), leftmargin=2em]
    \item For each mixture $k=1,...,K$, sample $M_{sub}$ trajectories $\{\bm{Y}_{k,m}\}_{m=1}^{M_{sub}}$. Each $\bm{Y}_{k,m}[t,:] \sim \mathcal{N}(\bm{\mu}_k[t,:], \text{diag}(\bm{\sigma}_k^2[t,:]))$.
    \item Create candidate list $C = \{(\bm{Y}_{k,m}, \pi_k)\}$. Sort $C$ by $\pi_k$ descending.
    \item Initialize selected set $S_{NMS} = \emptyset$. While $|S_{NMS}| < N_{samples}$ and $C \neq \emptyset$:
        \begin{enumerate}[label=(\roman*), leftmargin=1.5em]
            \item Pop best trajectory $\bm{Y}^*$ from $C$. Add $\bm{Y}^*$ to $S_{NMS}$.
            \item For each remaining $\bm{Y}_{cand} \in C$: if $ || \bm{Y}_{cand}[T_{fut},:2] - \bm{Y}^*[T_{fut},:2] ||_2 \le d_{NMS}$, remove $\bm{Y}_{cand}$ from $C$.
        \end{enumerate}
    \item If $|S_{NMS}| < N_{samples}$, pad $S_{NMS}$ by duplicating its highest probability trajectories.
\end{enumerate}

\subsection{Trajectory Smoothing (Model Predictive Control)}
For each of the $N_{samples}$ trajectories $\bm{X}_{ref,s}$ from NMS (after sector boundary).
\subsubsection{MPC Optimization}
Solve:
\[ \min_{\bm{U}_s = [\bm{u}_{s,0}, ..., \bm{u}_{s,T_{mpc}-1}]} J(\bm{U}_s) = \sum_{t=0}^{T_{mpc}-1} \left( (\bm{x}_{s,t+1} - \bm{x}_{ref,s,t+1})^T \bm{Q} (\bm{x}_{s,t+1} - \bm{x}_{ref,s,t+1}) + \bm{u}_{s,t}^T \bm{R} \bm{u}_{s,t} \right) \]
Subject to:
\begin{itemize}[leftmargin=1.5em]
    \item $\bm{x}_{s,t+1} = f(\bm{x}_{s,t}, \bm{u}_{s,t})$ (Ackermann dynamics as in previous response).
    \item $\bm{x}_{s,0} = \bm{x}_{init}$ (current ego state).
    \item $\bm{u}_{lower} \le \bm{u}_{s,t} \le \bm{u}_{upper}$.
\end{itemize}
This is solved using an SLSQP algorithm. The resulting state sequence is $\bm{X}_{smooth,s}$.

\subsection{Loss Calculation (Training)}
Let $\bm{Y}_{smooth,s,b,t}$ be the $(x,y,\psi)$ components of $\bm{X}_{smooth,s}$ for batch $b$ at time $t$.
\begin{itemize}[leftmargin=*]
    \item Position Loss: $\mathcal{L}_{pos} = \frac{1}{N_s B T_f} \sum_{s,b,t} || \bm{Y}_{smooth,s,b,t,:2} - \bm{Y}_{target,b,t,:2} ||_1$.
    \item Yaw Loss: $\mathcal{L}_{yaw} = \frac{1}{N_s B T_f} \sum_{s,b,t} | Y_{smooth,s,b,t,2} - Y_{target,b,t,2} |_1$.
    \item Uncertainty Loss: $\mathcal{L}_{unc} = - \frac{1}{K B T_f d_{act}} \sum_{k,b,t,d} \log(\sigma_{k,b,t,d}^2 + \epsilon_{small})$.
    \item KL Divergence Loss: For batch item $b$, $P_b = \text{Categorical}(\bm{\pi}_b)$, $Q_0 = \text{UniformCategorical}(K)$.
    \[ \mathcal{L}_{KL} = \frac{1}{B} \sum_{b=1}^{B} \sum_{k=1}^{K} \pi_{b,k} \log\left(\frac{\pi_{b,k}}{1/K}\right) \]
    \item Total Loss: $\mathcal{L}_{total} = \mathcal{L}_{pos} + w_{yaw} \mathcal{L}_{yaw} + w_{unc} \mathcal{L}_{unc} + w_{KL} \mathcal{L}_{KL}$.
\end{itemize}
Parameters are updated via $\nabla_{\theta} \mathcal{L}_{total}$ using AdamW.

\section{A Simple Simulated Scenario}

We design a simulated experiment to vividly show our method's ability to reduce compounding error and generate diverse, plausible trajectories. In the experiment, we use synthetic data from an idealized scenario where a Self-Driving Vehicle (SDV) moves along a circular ring road network. The SDV maintains a fixed speed of 1 $m/s$, and data is sampled at 1 Hz. The circular lane is implicitly defined by the agent's motion and its relation to the nearest point on the ideal circular path.

Our proposed method implementation:
1.  **Historical Trajectory Features**: The SDV's past `$history\_frames$` (10) states, each consisting of position (x,y) and yaw, are transformed into an agent-centric coordinate system. This yields relative positions and relative yaws for these 10 historical points, which are then flattened.
2.  **Nearest Point Feature**: The 2D position of the point on the ideal circular path that is closest to the SDV's current (most recent historical) position is computed. This nearest point is then transformed into the agent-centric coordinate system.
The final input vector for the model is a concatenation of the flattened historical trajectory features (10 steps * 3 features/step = 30 features) and the 2 features from the relative nearest point, resulting in an input size of 32.

\begin{figure}[ht]
    \centering
    \subfloat[BC-perturb]{\includegraphics[width=0.49\textwidth]{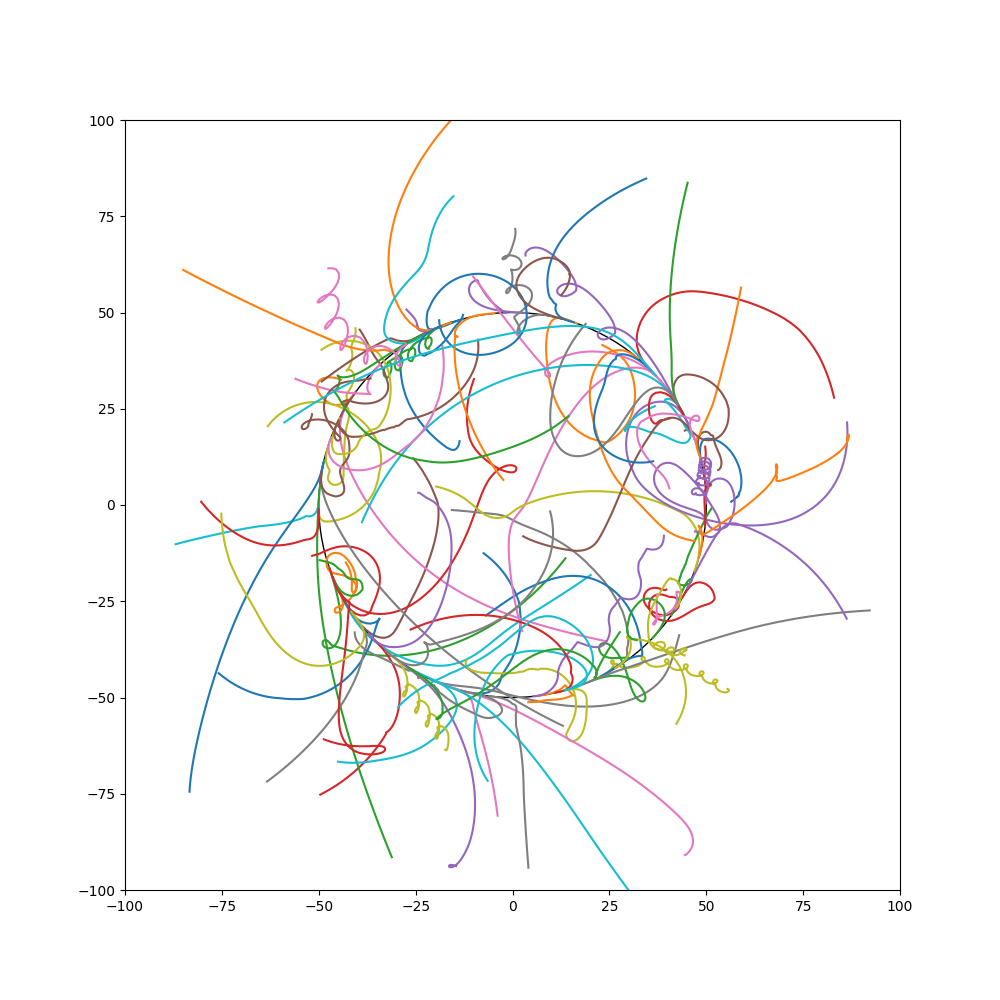}\label{subfig:bc}}
    \hfill
    \subfloat[UrbanDriver]{\includegraphics[width=0.49\textwidth]{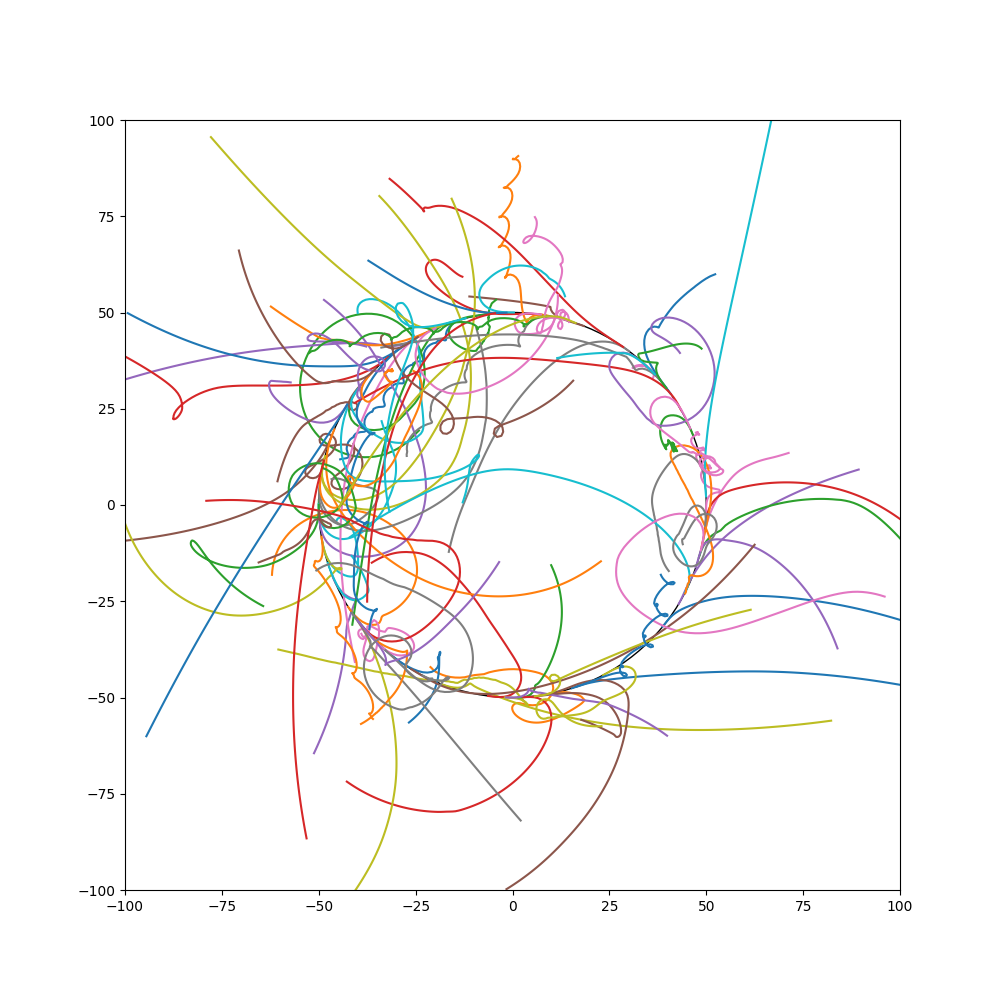}\label{subfig:urban}}
    \\
    \subfloat[CCIL]{\includegraphics[width=0.49\textwidth]{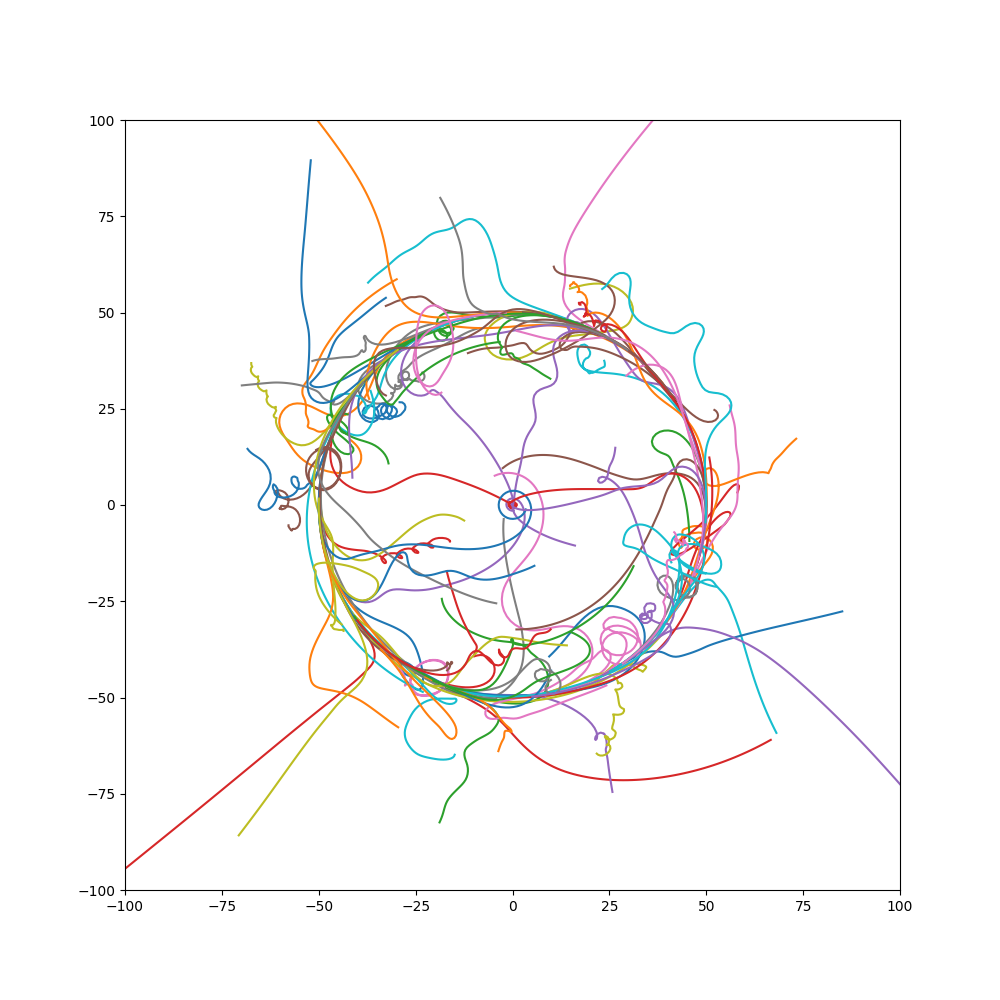}\label{subfig:ccil}}
    \hfill
    \subfloat[PhysVarMix]{\includegraphics[width=0.49\textwidth]{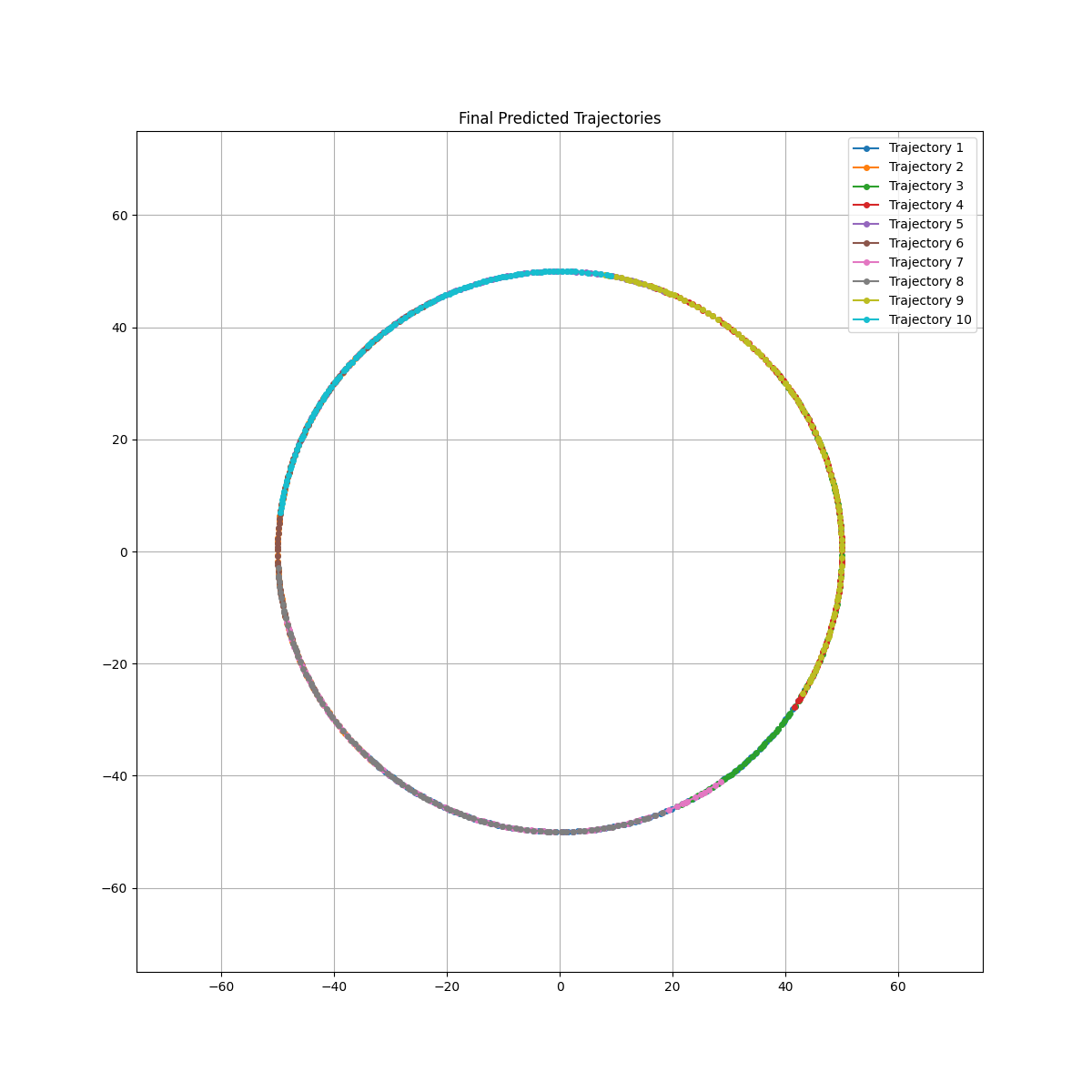}\label{subfig:phyvar}}
    \caption{We compare our methods with several baseline methods introduced above, including BC-perturb, CCIL, and UrbanDriver. A simulated scenario with closed-loop trajectories for 1000 steps' training}
    \label{fig:simu_my_method}
\end{figure}

Training is conducted for 1000 iterations. For evaluation, the trained policy is unrolled from random starting positions on a circle of radius 50 m for 100 time steps. The resulting closed-loop trajectories of our algorithm, as visualized in~\ref{fig:simu_my_method}, can be compared to those from baseline methods to demonstrate its effectiveness in maintaining on-road behavior and mitigating issues like covariate shift.

\paragraph{\textbf{The project is available from:}}

https://github.com/HaynesLi/PhysVarMix-Physics-Informed-Variational-Mixture-Model-for-Multi-Modal-Trajectory-Prediction


\bibliographystyle{abbrv}

\bibliography{ref}

\begin{thebibliography}{10}

\bibitem{bae2024language}
I.~Bae, J.~Lee, and H.-G. Jeon.
\newblock Can language beat numerical regression? language-based multimodal trajectory prediction, 2024.

\bibitem{bansal2019chauffeur}
M.~Bansal, A.~Krizhevsky, and A.~Ogale.
\newblock Chauffeurnet: Learning to drive by imitating the best and synthesizing the worst.
\newblock In {\em Proceedings of Robotics: Science and Systems (RSS)}. Robotics: Science and Systems Foundation, June 2019.

\bibitem{bishop1994mixture}
C.~M. Bishop.
\newblock Mixture density networks.
\newblock Technical report, Aston University, 1994.

\bibitem{caesar2021nuplan}
H.~Caesar, J.~Kabzan, K.~S. Tan, W.~K. Fong, E.~Wolff, A.~Lang, L.~Fletcher, O.~Beijbom, and S.~Omari.
\newblock nuplan: A closed-loop ml-based planning benchmark for autonomous vehicles.
\newblock In {\em 2021 IEEE/CVF Conference on Computer Vision and Pattern Recognition Workshops (CVPRW)}, pages 4237--4246. IEEE, June 2021.

\bibitem{chai2019multipath}
Y.~Chai, B.~Sapp, M.~Bansal, and D.~Anguelov.
\newblock Multipath: Multiple probabilistic anchor trajectory hypotheses for behavior prediction.
\newblock {\em arXiv preprint arXiv:1910.05449}, 2019.

\bibitem{chen2022learning}
J.~Chen, H.~Zhang, Y.~Li, and J.~Sun.
\newblock Learning map representations for autonomous driving: A survey.
\newblock {\em IEEE Transactions on Intelligent Vehicles}, 7(4):891--905, December 2022.

\bibitem{cui2018multimodal}
H.~Cui, V.~Radosavljevic, F.-C. Chou, T.-H. Lin, T.~Nguyen, T.-K. Huang, J.~Schneider, and N.~Djuric.
\newblock Multimodal trajectory predictions for autonomous driving using deep convolutional networks.
\newblock {\em arXiv preprint arXiv:1809.10732}, 2018.

\bibitem{falcone2007predictive}
P.~Falcone, F.~Borrelli, J.~Asgari, H.~E. Tseng, and D.~Hrovat.
\newblock Predictive active steering control for autonomous vehicle systems.
\newblock {\em IEEE Transactions on Control Systems Technology}, 15(3):566--580, May 2007.

\bibitem{fujimoto2021minimalist}
S.~Fujimoto and S.~S. Gu.
\newblock A minimalist approach to offline reinforcement learning.
\newblock In {\em Advances in Neural Information Processing Systems 34 (NeurIPS 2021)}, volume~34, pages 20132--20145. Curran Associates, Inc., December 2021.
\newblock Available at: https://papers.nips.cc/paper/2021/file/a66f42a6b49c6d656f0a76d4dfdf636c-Paper.pdf.

\bibitem{gao2024intentiondiffusion}
X.~Gao, X.~Gao, J.~Zhao, H.~Wu, Z.~Liu, and M.~Li.
\newblock Intention-aware denoising diffusion model for trajectory prediction, 2024.

\bibitem{gilles2021probabilistic}
T.~Gilles, S.~Sabatini, D.~Tsishkou, B.~Stanciulescu, and F.~Moutarde.
\newblock Probabilistic multimodal trajectory prediction with lane attention for autonomous vehicles.
\newblock In {\em 2021 IEEE/RSJ International Conference on Intelligent Robots and Systems (IROS)}, pages 1428--1434. IEEE, September 2021.

\bibitem{girgis2022latent}
R.~Girgis, F.~Golemo, F.~Codevilla, M.~Weiss, S.~D'Souza, S.~E. Kahou, F.~Heide, and C.~Pal.
\newblock Latent variable sequential set transformers for joint multi-agent motion prediction.
\newblock In {\em International Conference on Learning Representations}, 2022.

\bibitem{gomes2022review}
I.~Gomes and D.~Wolf.
\newblock A review on intention-aware and interaction-aware trajectory prediction for autonomous vehicles.
\newblock In {\em 2022 Latin American Robotics Symposium (LARS), 2022 Brazilian Symposium on Robotics (SBR), and 2022 Workshop on Robotics in Education (WRE)}, pages 84--89. IEEE, October 2022.

\bibitem{guo2023ccilcontextconditionedimitationlearning}
K.~Guo, W.~Jing, J.~Chen, and J.~Pan.
\newblock Ccil: Context-conditioned imitation learning for urban driving, 2023.

\bibitem{hari2024navigation}
A.~Hari, Z.~Liu, and R.~Mangharam.
\newblock Navigation under uncertainty: trajectory prediction and occlusion reasoning with switching dynamical systems, 2024.

\bibitem{he2016deep}
K.~He, X.~Zhang, S.~Ren, and J.~Sun.
\newblock Deep residual learning for image recognition.
\newblock In {\em 2016 IEEE Conference on Computer Vision and Pattern Recognition (CVPR)}, pages 770--778. IEEE, June 2016.

\bibitem{hochreiter1997long}
S.~Hochreiter and J.~Schmidhuber.
\newblock Long short-term memory.
\newblock {\em Neural Computation}, 9(8):1735--1780, 1997.

\bibitem{houston2020one}
J.~Houston, G.~Zuidhof, L.~Bergamini, Y.~Ye, L.~Chen, A.~Jain, S.~Omari, V.~Iglovikov, and P.~Ondruska.
\newblock \href{https://proceedings.mlr.press/v155/houston21a/houston21a.pdf}{One thousand and one hours: Self-driving motion prediction dataset}.
\newblock In {\em CoRL}, 2020.

\bibitem{huang2022survey}
Y.~Huang, Y.~Fan, X.~Yang, B.~Zhao, R.~Yuan, R.~Liu, Y.~Zhang, and P.~Li.
\newblock A survey on trajectory-prediction methods for autonomous driving.
\newblock {\em IEEE Transactions on Intelligent Vehicles}, 7(3):652--674, September 2022.

\bibitem{jeong2019motion}
Y.~Jeong, S.~Lee, and K.~Yi.
\newblock Motion prediction of surrounding vehicles using deep neural networks for autonomous driving.
\newblock In {\em 2019 IEEE Intelligent Vehicles Symposium (IV)}, pages 1928--1933. IEEE, June 2019.

\bibitem{karunakaran2023efficient}
D.~Karunakaran, J.~Yang, S.~Fang, and W.~Wang.
\newblock Efficient prediction and uncertainty propagation for motion planning with multimodal behavior distribution.
\newblock In {\em 2023 IEEE 26th International Conference on Intelligent Transportation Systems (ITSC)}, pages 3378--3385. IEEE, September 2023.

\bibitem{kim2024singulartrajectory}
S.~Kim, J.~Lee, and H.-G. Jeon.
\newblock Singulartrajectory: Universal trajectory predictor using diffusion model, 2024.

\bibitem{lee2017desire}
N.~Lee, W.~Choi, P.~Vernaza, C.~B. Choy, P.~H. Torr, and M.~Chandraker.
\newblock Desire: Distant future prediction in dynamic scenes with interacting agents.
\newblock In {\em Proceedings of the IEEE Conference on Computer Vision and Pattern Recognition}, pages 300--309, 2017.

\bibitem{lefevre2014survey}
S.~Lef{\`e}vre, D.~Vasquez, and C.~Laugier.
\newblock A survey on motion prediction and risk assessment for intelligent vehicles.
\newblock {\em Robomech Journal}, 1(1):1--14, 2014.

\bibitem{leng2024trajllm}
Z.~Leng, H.~Wang, S.~Han, Y.~Xing, K.~Tang, J.~Zhu, and Z.~Xing.
\newblock Traj-llm: A new exploration for empowering trajectory prediction with pre-trained large language models, 2024.

\bibitem{leon2021review}
F.~Leon and M.~Gavrilescu.
\newblock A review of tracking and trajectory prediction methods for autonomous driving.
\newblock {\em Mathematics}, 9(6):660, March 2021.

\bibitem{li2021spatio}
J.~Li, X.~Tao, Y.~Guo, and J.~Lu.
\newblock Spatio-temporal trajectory prediction with graph neural networks for autonomous driving.
\newblock {\em IEEE Transactions on Vehicular Technology}, 70(8):7666--7678, August 2021.

\bibitem{li2025hybrid}
Z.~Li and D.~Pfoser.
\newblock Hybrid machine learning model with a constrained action space for trajectory prediction, 2025.

\bibitem{liang2020learning}
M.~Liang, B.~Yang, R.~Hu, Y.~Chen, R.~Liao, S.~Feng, and R.~Urtasun.
\newblock Learning lane graph representations for motion forecasting.
\newblock In {\em European Conference on Computer Vision}, pages 541--556. Springer, 2020.

\bibitem{mercat2020multi}
J.~Mercat, L.~Bartolo, J.~Pettré, and N.~L. Priol.
\newblock Multi-head attention for multi-modal trajectory prediction in autonomous driving.
\newblock In {\em 2020 IEEE International Conference on Robotics and Automation (ICRA)}, pages 5688--5694. IEEE, May 2020.

\bibitem{mozaffari2021deep}
S.~Mozaffari, O.~Y. Al-Jarrah, M.~Dianati, E.~Moulas, and S.~Fallah.
\newblock Deep learning-based vehicle behavior prediction for autonomous driving applications: A review.
\newblock {\em IEEE Transactions on Intelligent Transportation Systems}, 22(1):33--47, January 2021.

\bibitem{mozaffari2022multimodal}
S.~Mozaffari, E.~Arnold, M.~Dianati, and S.~Fallah.
\newblock Multimodal maneuver prediction for autonomous vehicles using vehicular sensor and v2x data fusion.
\newblock {\em IEEE Transactions on Intelligent Transportation Systems}, 23(12):23757--23770, December 2022.

\bibitem{nayakanti2022wayformer}
N.~Nayakanti, R.~Al-Rfou, A.~Zhou, K.~Goel, K.~S. Refaat, and B.~Sapp.
\newblock Wayformer: Motion forecasting via simple \& efficient attention networks.
\newblock {\em arXiv preprint arXiv:2207.05844}, 2022.

\bibitem{ngiam2021scene}
J.~Ngiam, B.~Caine, W.~Han, B.~Yang, Y.~Chai, P.~Lu, X.~Peng, V.~Vasudevan, X.~Zhou, A.~Chouard, et~al.
\newblock Scene transformer: A unified architecture for predicting multiple agent trajectories.
\newblock {\em arXiv preprint arXiv:2106.08417}, 2021.

\bibitem{ren2015faster}
S.~Ren, K.~He, R.~Girshick, and J.~Sun.
\newblock Faster r-cnn: Towards real-time object detection with region proposal networks.
\newblock In {\em Advances in neural information processing systems}, pages 91--99, 2015.

\bibitem{salzmann2020trajectron}
T.~Salzmann, B.~Ivanovic, P.~Chakravarty, and M.~Pavone.
\newblock Trajectron++: Dynamically-feasible trajectory forecasting with heterogeneous data.
\newblock In {\em 2020 European Conference on Computer Vision (ECCV)}, volume 12363 of {\em Lecture Notes in Computer Science}, pages 683--700. Springer, August 2020.

\bibitem{scheel2021urban}
O.~Scheel, L.~Schwarz, B.~Knopf, S.~Casas, and R.~Urtasun.
\newblock Urbandriver: Learning driver behaviors with multimodal imitation learning.
\newblock {\em arXiv preprint arXiv:2109.13333}, September 2021.
\newblock Available at: https://arxiv.org/abs/2109.13333.

\bibitem{schreier2016integrated}
M.~Schreier, V.~Willert, and J.~Adamy.
\newblock An integrated approach to maneuver-based trajectory prediction and criticality assessment in arbitrary road environments.
\newblock In {\em IEEE Transactions on Intelligent Transportation Systems}, volume~17, pages 2751--2766. IEEE, October 2016.

\bibitem{sohn2015learning}
K.~Sohn, H.~Lee, and X.~Yan.
\newblock Learning structured output representation using deep conditional generative models.
\newblock In {\em Advances in neural information processing systems}, pages 3483--3491, 2015.

\bibitem{sriramulu2024multitransmotion}
A.~Sriramulu, N.~Fourrier, and C.~Bergmeir.
\newblock Multi-transmotion: Pre-trained model for human motion prediction, 2024.

\bibitem{tumu2023physics}
R.~Tumu, L.~Lindemann, T.~Nghiem, and R.~Mangharam.
\newblock Physics constrained motion prediction with uncertainty quantification, 2023.

\bibitem{ulbrich2015towards}
S.~Ulbrich, T.~Menzel, A.~Reschka, F.~Schuldt, and M.~Maurer.
\newblock Towards a functional system architecture for automated vehicles.
\newblock In {\em 2015 IEEE 18th International Conference on Intelligent Transportation Systems (ITSC)}, pages 1420--1425. IEEE, September 2015.

\bibitem{vaswani2017attention}
A.~Vaswani, N.~Shazeer, N.~Parmar, J.~Uszkoreit, L.~Jones, A.~N. Gomez, {\L}.~Kaiser, and I.~Polosukhin.
\newblock Attention is all you need.
\newblock In {\em Advances in neural information processing systems}, pages 5998--6008, 2017.

\bibitem{wang2023survey}
Y.~Wang, H.~Wu, C.~Liu, and C.~Lv.
\newblock A survey on deep learning-based autonomous driving: Methods and challenges.
\newblock {\em IEEE Transactions on Intelligent Transportation Systems}, 24(5):4765--4784, May 2023.

\bibitem{xia2024robusttrajectory}
J.~Xia, C.~Xu, Y.~Yan, Z.~Fan, X.~Gong, W.~Huang, and F.~Yu.
\newblock Towards robust trajectory representations: Isolating environmental confounders with causal learning, 2024.

\bibitem{yuan2020dlow}
Y.~Yuan and K.~Kitani.
\newblock Dlow: Diversifying latent flows for diverse human motion prediction.
\newblock In {\em European Conference on Computer Vision}, pages 346--364. Springer, 2020.

\bibitem{zhao2021review}
X.~Zhao, H.~Xiang, W.~Yang, C.~Feng, and J.~Wang.
\newblock A review of deep learning-based vehicle motion prediction for autonomous driving.
\newblock {\em IEEE Transactions on Intelligent Transportation Systems}, 22(9):5667--5684, September 2021.

\bibitem{zhou2024smartpretrain}
H.~Zhou, H.~Liu, Z.~Zhang, Y.~Xu, J.~Yang, and J.~Zhang.
\newblock Smartpretrain: Model-agnostic and dataset-agnostic representation learning for motion prediction, 2024.

\bibitem{zhou2023vector}
H.~Zhou, Y.~Xing, and C.~Lv.
\newblock Vectorized map representation for autonomous driving: A review.
\newblock {\em IEEE Transactions on Intelligent Transportation Systems}, 24(7):6789--6804, July 2023.

\end{thebibliography}

\end{document}